\documentclass{bmvc2k}

\usepackage{textcomp}
\usepackage{verbatim}
\usepackage{amsmath}
\usepackage{capt-of}
\usepackage{enumitem}

\newcommand{\beginsupplement}{%
        \setcounter{table}{0}
        \renewcommand{\thetable}{S\arabic{table}}%
        \setcounter{figure}{0}
        \renewcommand{\thefigure}{S\arabic{figure}}%
     }

\title{Text and Style Conditioned GAN for Generation of Offline Handwriting Lines}

\addauthor{Brian Davis}{briandavis@byu.net}{1}
\addauthor{Chris Tensmeyer}{tensmeye@adobe.com}{2}
\addauthor{Brian Price}{bprice@adobe.com}{2}
\addauthor{Curtis Wigington}{wigingto@adobe.com}{2}
\addauthor{Bryan Morse}{morse@cs.byu.edu}{1}
\addauthor{Rajiv Jain}{rajijain@adobe.com}{2}

\addinstitution{
 Brigham Young University\\
 Provo, Utah, USA
}
\addinstitution{
 Adobe Research\\
 San Jose, California, USA
}

\runninghead{Davis, et al.}{Text and Style Conditioned GAN for Offline-Handwriting}

\begin{document}

\maketitle

\begin{abstract}
This paper presents a GAN for generating images of handwritten lines 
conditioned on arbitrary text and latent style vectors. 
Unlike prior work, which produce stroke points or single-word images, this model generates entire lines of offline handwriting.
The model produces variable-sized images by using style vectors to determine character widths. 
A generator network is trained with GAN and autoencoder techniques to learn style, and uses a pre-trained handwriting recognition network to induce legibility.
A study using human evaluators demonstrates that the model produces images that appear to be written by a human.
After training, the encoder network can extract a style vector from an image, allowing images in a similar style to be generated, but with arbitrary text.
\end{abstract}


\section{Introduction}

\begin{figure}[t]
\centering
\vspace{0.01in}
\includegraphics[width=0.70\textwidth]{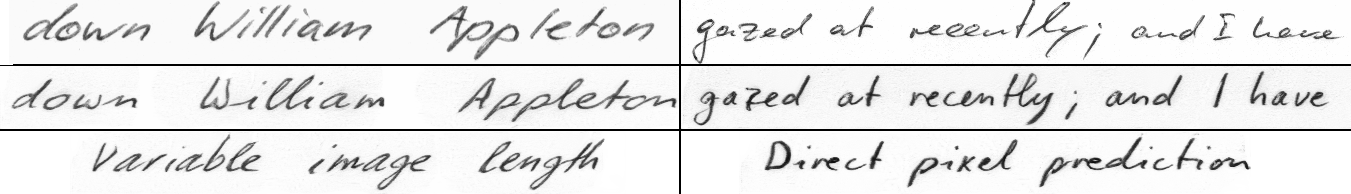}
\vspace{-0.1in}
\caption{Examples of our model mimicking two authors' style. Top: original authors' writing. Middle: reconstructions using our model. Bottom: novel text using the same style.
\vspace{-0.1in}
}
\label{fig:top}
\end{figure}

In this work, we generate images of lines of handwriting, conditioned on the desired text and a latent style vector. 
Handwriting is an expressive and unique form of communication that is often considered more intimate than typed text. 
Generating images that mimic an author's style would allow people to generate their own handwriting from typed text. 
While a convenience for many, this is particularly valuable to those with physical disabilities that hinder or prevent them from writing.
Our results achieve human plausibility and begin to approximate the style of example handwriting images, as seen in Fig.~\ref{fig:top}.

Handwritten image generation can also provide additional data to train more accurate general handwriting recognition models~\cite{AlonsoGeneration}.
Generating a large number of images in the style of each user of an application (e.g., scanner) could allow us to train personalized single-author recognition models.  
Personalized models tend to be more accurate for their target author's writing than a general recognition model. 


Several previous approaches have framed handwriting generation as stroke prediction~\cite{Graves2013GeneratingSW,Nakano2019NeuralPA,Mellor2019UnsupervisedDA}, i.e.,~modeling how a pen moves on paper. 
Training these methods often requires \emph{online} handwriting data, captured by writing on a digital device.
Without post-processing, such methods do not model ink textures that result from writing on physical media.
In contrast, our method uses widely available \emph{offline} handwriting data, i.e.,~images of the physical media.
We frame handwriting generation as conditional image generation, directly learning from and predicting pixels~\cite{AlonsoGeneration,Kang2020GANwritingCG,scrabblegan}. 

Our approach achieves realism by combining Generative Adversarial Networks (GAN)~\cite{goodfellow2014generative} and autoencoder methods~\cite{vae-auto-larsen16} with an auxiliary loss to achieve legibility~\cite{AlonsoGeneration}.
Our encoder can map an example image into a latent style space, and then the model can produce images in a similar style, either reconstructing the original or using arbitrary text.

The width of an image of real handwriting depends on the text and writing style.
Instead of fixing our model's output size or heuristically determining the output width from the input text, we use a deep network to predict the horizontal layout of the characters. 
Our model achieves state-of-the-art visual quality for offline handwriting generation and, unlike prior methods \cite{AlonsoGeneration,Kang2020GANwritingCG}, generates entire lines of handwriting conditioned on arbitrarily long text.

It is challenging to train a network using many (sometimes competing) loss functions that yield gradients that differ in size by orders of magnitude.
To overcome this, we propose an improvement upon the gradient balancing technique of~\cite{AlonsoGeneration}.

There are potential ethical concerns due to nefarious uses of this technology, e.g., low-skill convincing forgery. 
However, we believe this concern is minor as the method is not targeted at imitating signatures and can only produce digital images, not physical documents.

Our primary contributions are: (1) a method combining GANs and autoencoders to train a handwriting generator on offline handwriting images to produce realistic handwritten images that mimic example image styles; (2) a model that generates variable-length handwritten line images from arbitrary length text and style; and (3) improved multi-loss training through gradient balancing, allowing disparate losses to be used together more easily.
Our code is available at \url{https://github.com/herobd/handwriting_line_generation}.

\section{Prior Work}

Conditional image generation methods take as input a description of the desired output image.
Descriptions used in prior work include semantic layout masks~\cite{guagan,Chen_2017_ICCV,Isola_2017_CVPR,Wang_2018_CVPR}, sketches~\cite{Sketch2Photo,Isola_2017_CVPR,Wang_2018_CVPR}, image and desired attribute~\cite{choi2018stargan}, image classes~\cite{mirza2014conditional}, key-words~\cite{photosynthesisJohnson} and natural language descriptions~\cite{Zhang_2017_ICCV,zhang2018stackgan++}.
Many recent image generation approaches employ GANs~\cite{goodfellow2014generative}, where the generator 
produces samples to fool a discriminator 
that attempts to classify images as real or generated.
In our work, we employ a GAN and condition the output on both a target text and a latent style vector.

Recent GANs model the content and style of an image.
StyleGAN~\cite{stylegan} and the improved StyleGAN2~\cite{karras2019analyzing} learn a mapping from random noise vectors to style vectors that influence style by controlling the mean and magnitude of the generator network feature maps via AdaIN layers.
MUNIT~\cite{huang2018multimodal} translates images from one domain to another by learning autoencoders that encode the input image as separate, latent content and style vectors.
FUNIT~\cite{funit} builds on MUNIT to allow the target class to be unseen during training and instead be specified by a handful of images at test time.
We similarly employ an autoencoder that reconstructs images from style and content, but our encoder only needs to extract the style vector.
In contrast to both these methods, our content description is a variable-length text string instead of a fixed-sized latent vector, and we produce variable-width images based on the combination of the style vector and content.

Graves's well-known \emph{online} handwriting generation LSTMs~\cite{Graves2013GeneratingSW} predict future stroke points from previous stroke points and an input text.
In contrast, we directly generate an image of \emph{offline} handwriting that includes realistic ink textures.
The authors of~\cite{Aksan:2018:DeepWriting} use RNNs to perform online generation and explicitly model content and style separately.
In~\cite{Ji2019GenerativeAN}, a GAN framework is used to train the generator of~\cite{Graves2013GeneratingSW}.

Alonso et al.~\cite{AlonsoGeneration} proposed an offline handwriting generator for isolated, fixed-size word images. It is trained using offline handwriting images. Their generator conditions on a fixed-size RNN-learned text embedding and random style vector.
They train the generator in GAN fashion produce word images that fool a discriminator, but also include a loss to encourage legible text according to a jointly-trained handwriting recognition model.
Specifically, the generator is updated to minimize the CTC loss~\cite{CTC} between the output of the recognition model and the input text.
We similarly employ a recognition model to improve legibility, but we instead use a pre-trained feedforward network.
In contrast, we produce higher quality handwritten line images from arbitrarily long text (instead of just words), and we can extract styles from existing images to generate similarly styled images.


Contemporary work~\cite{Kang2020GANwritingCG} improves upon~\cite{AlonsoGeneration} by extracting style from a set of 15 word images from a single author.
Generated images are fed to a writer classifier, and to learn writer styles, the generator is updated to fool this classifier.
The generated word images are realistic, but don't recreate style perfectly.
Recently, ScrabbleGAN~\cite{scrabblegan} improved upon~\cite{AlonsoGeneration} by making the generated image width proportional to the input text length.
In contrast, our model learns the output width based on the provided style and the input text.


\section{Method}\label{sec:method}

We view the handwriting generation process as having three inputs: content, style, and noise. 
Content is the desired text. 
Style is the unique way a writer forms characters using a particular physical medium and writing instrument.
Noise is the natural variation of individual handwriting, even when writing the same content in the same style.  

\begin{figure}[t!]
\centering
\includegraphics[width=0.95\textwidth]{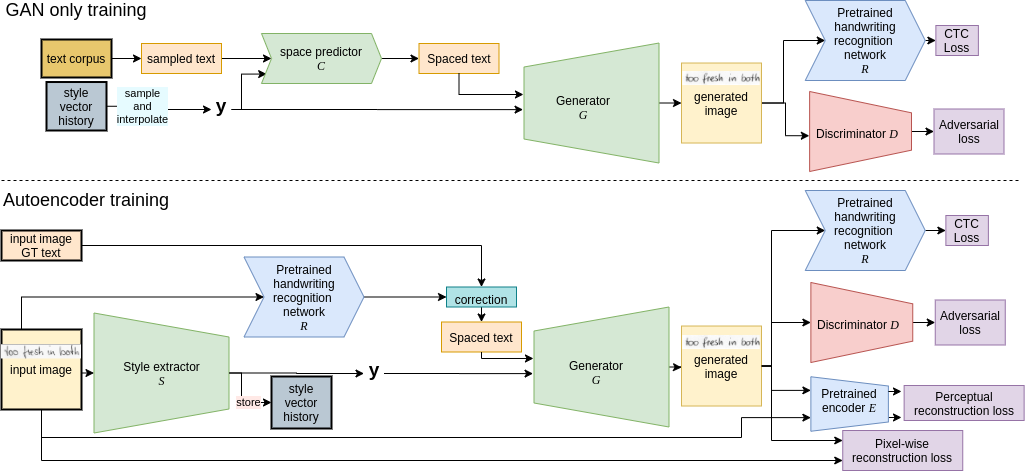}
\vspace{-0.1in}
\caption{Overview of our method. To generate an image, we take input text and style vector \textbf{y}. Text is spaced by spacer $C$ using \textbf{y}. This spaced text and \textbf{y} are passed to the generator $G$. We use both fully GAN training steps (top) and autoencoder based training steps (bottom).
\vspace{-0.15in}
}
\label{fig:full}
\end{figure}

We train our model with GAN, reconstruction, perceptual, and text recognition losses.
Fig.~\ref{fig:full} shows an overview of our training process, including six networks:
(1)~a generator network $G$ to produce images from spaced text, a style vector, and noise.
(2)~a style extractor network $S$, that produces a style vector from an image and the recognition predictions; 
(3)~a spacing network $C$, which predicts the horizontal text spacing based on the style vector; 
(4)~a patch-based convolutional discriminator $D$; 
(5)~a pretrained handwriting recognition network $R$ to encourage image legibility and correct content; 
and 
(6)~a pretrained encoder $E$, to compute a perceptual loss.
    


During training the model learns to mimic style as $S$ and $G$ act as an encoder and decoder in an autoencoder, with $E$ helping supply the reconstruction loss. $G$ requires the input text to have spacing information (spaced text), which is extracted from the target image using $R$ and the ground truth text (Sec.~\ref{sec:spaced_text}). $G$ and $D$ act as a GAN, supplying the model an adversarial loss for realism. $R$ allows a handwriting recognition loss to supervise the legibility of generated images. Sampling a  style vector and predicting spaced text using $C$ allows the model produce novel images. $C$ is supervised using styles predicted by $S$ (not pictured in Fig~\ref{fig:full}).

We now present details for each part of our model and its training.
Full architectural diagrams for $G$, $S$, $C$, $D$, $E$, and $R$ are provided in Supplementary Material (\ref{sec:model_supp}).

\subsection{Generator $G$ and Discriminator $D$}

$G$ is based on StyleGAN~\cite{stylegan} but differs in architecture and receives the 1D spaced text as input with the style vector concatenated at each spatial position. 
Spaced text (Sec.~\ref{sec:spaced_text}) is a one-hot encoding of the target text with additional blank characters and repeated characters, which encode the spacing information. 
This informs  horizontal character placement and was key to training the model successfully. 
The network blocks consist of a convolutional layer, additive noise, ReLU activation, and AdaIN~\cite{stylegan}, which uses the style vector to determine feature map statistics.
To increase resolution, we use nearest-neighbor upsampling followed by a convolution and blurring operation.
Most upsampling is in only the vertical dimension because the spaced text input is already wide.

Our discriminator $D$ must be able handle variable sized inputs, so it is a fully convolutional, multi-resolution patch-based discriminator that we train with a hinge loss.


\subsection{Style Extractor $S$}



\begin{figure}[t!]
    \centering
    \begin{minipage}{0.66\textwidth}
        \centering
        \includegraphics[width=0.99\textwidth]{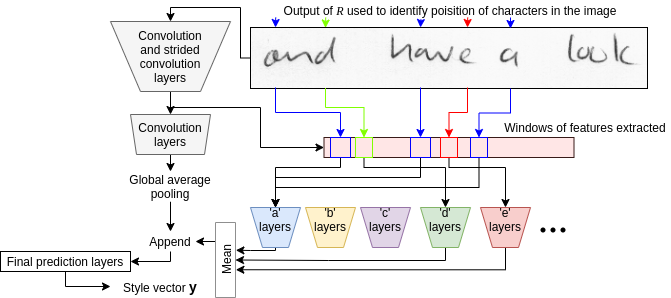} 
        \vspace{-.1in}
        \caption{Style Extractor $S$ architecture with character-specific heads to process feature windows from detected character locations.
        \vspace{-0.05in}
}
\label{fig:style_ext}
    \end{minipage}
    \hspace{0.02\textwidth}
    \begin{minipage}{0.3\textwidth}
        \centering
        \includegraphics[width=0.81\textwidth]{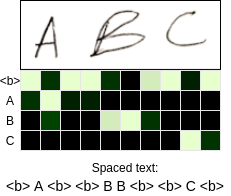} 
        \vspace{-.1in}
        \caption{
        Example output of $R$ (lighter is higher prob.) and corresponding spaced text.
        \vspace{-0.15in}
}
\label{fig:spaced_text}
    \end{minipage}
\end{figure}

$S$ inputs the image and the output of $R$ on the image to produce a style vector (Fig~\ref{fig:style_ext}).
First, it uses a convolutional network to extract a 1D (horizontal) sequence of features.
Then the recognition result (Fig.~\ref{fig:spaced_text}) is used to roughly localize each recognized character in the feature sequence.
For each predicted character, we crop the feature sequence with a window size of 5 (roughly 40 pixels, an area slightly larger than most characters in the IAM dataset) centered on the character and then pass each window through character-specific layers to extract character features.
Features from all instances of all characters are averaged, weighted by $R$'s predicted confidence for each instance, giving a final character feature vector.

To obtain global style features, we pass the entire feature sequence through 1D convolutional layers and perform global average pooling.
This is appended to the character feature vector, from which fully connected layers predict the final style vector of dimension 128.

\subsection{Spaced Text and Spacing Network $C$}
\label{sec:spaced_text}

We found spaced text essential for training with a reconstruction loss. 
Without it, $G$ has difficulty achieving horizontal alignment with the input image and fails to train. 
Spaced text can be derived for a particular image from the output of $R$ or predicted directly by $C$ for a novel style.

Width and spacing are encoded using repeated characters and blank symbols {\tt <b>} (Fig.~\ref{fig:spaced_text}).
\emph{Dataset spaced text} is obtained by taking the predicted character at each horizontal position from the output of $R$ on a dataset image (Fig.~\ref{fig:spaced_text}), keeping blanks and repeated characters (artifacts typically removed when decoding the output of a CTC trained model).
We correct recognition errors in the dataset spaced text using the ground truth text.

$C$ is a 1D convolutional network that consumes one-hot encoded target text with the style vector concatenated to each position.
For each character $c_i$, C predicts how many blanks precedes $c_i$ and how many times $c_i$ is repeated in the spaced text.
Multiple blanks are then appended to the output.
$C$ is trained to imitate the dataset spaced text using a MSE loss.

\subsection{Handwriting Recognition Network $R$}

$R$ is a pretrained handwriting recognition network that encourages generated images to (legibly) contain the specified text by applying the CTC loss~\cite{CTC}.
$R$'s weights are frozen so the gradient merely flows through $R$ to supervise $G$.

While state-of-the-art handwriting recognition methods \cite{startfollowread,graves2009offline,Puigcerver} use CNN-RNNs, we obtained better results with $R$ as a fully convolutional network based on~\cite{startfollowread}. 
RNNs have arbitrarily large context windows and may predict characters based on linguistic context instead of visual character shapes.
In contrast, $R$ only uses local visual features for character predictions and therefore provides better feedback for generating characters.

$R$ is pretrained with CTC loss and warp grid data augmentation~\cite{augmentation}, which results in overall better generated images, but training the model on warped images causes some artifacts in the absence of the reconstruction loss (e.g., the first ablated model in Fig.~\ref{fig:ablation}).


\subsection{Encoder Network $E$}\label{sec:encoder}

$E$ provides features for our perceptual loss \cite{perceptual}.
Traditional methods for computing perceptual loss use features from pretrained image classification networks~\cite{johnson2016perceptual}. 
However, images of handwriting are different from natural images, so we choose a different approach. 
$E$~is a fully convolutional network that collapses the image to a one-dimensional feature series capturing visual and semantic features. 
$E$ is trained  both as an autoencoder with a decoder and L1 reconstruction loss, and as a handwriting recognizing network with CTC loss. 

\subsection{Training/Losses}

Our generation has three objectives: legible handwriting matching the target text, realistic handwriting that appears to be a human's, and handwriting style that mimics an example image. 
Each of these is achieved primarily through the respective losses: CTC loss back-propagating through $R$, adversarial loss, and reconstruction losses (pixel and perceptual). 
Additionally, MSE is used to train the spacing network~$C$, and hinge loss is used to train~$D$.



When using multiple loss terms, balancing them is crucial. 
We do so by improving the gradient-balancing method of~\cite{AlonsoGeneration}.
Without balancing, training failed due to exploding gradients or failed to converge.
Stable hyper-parameters possibly exist, but gradient balancing easily solved the problem. 
We balance gradients from CTC, adversarial, and reconstruction losses. 
The two reconstruction losses have equal weight and are summed.

In \cite{AlonsoGeneration} the CTC loss gradient is normalized to have the same mean and standard deviation as the adversarial loss gradient. 
However, this does not preserve the sign of the CTC gradient, so we instead normalize the gradients to have the same mean magnitude (per layer). This additionally allows balancing multiple gradients.
Totally equal contributions may not be desirable and can be adjusted by multiplicative weights on each gradient after normalization. 
We always use the gradient magnitude of the reconstruction loss for gradient normalization. 

To reduce memory requirements, some training steps store only gradients (for later balancing) and others update the parameters. 
Our curriculum uses the following steps:
\begin{enumerate}[topsep=1pt,itemsep=-1ex,partopsep=1ex,parsep=1ex]
    \item \textbf{Spacing:} This is skipped on every other round through the curriculum. A style is extracted from two dataset images by the same author and $C$ predicts the spacing. The MSE loss between the prediction and dataset spaced text updates both $C$ and $S$.
    
    \item  \textbf{Discriminator:} To update $D$ we sample novel styles by interpolating/extrapolating styles sampled from a running window history of the 100 most recently extracted styles (during Spacing or Autoencoder steps). 
    Extrapolation is kept within 0.5 of the distance between the two styles and is sampled uniformly from that range. 
    
    \item \textbf{GAN-only:} This follows standard GAN training while including the handwriting recognition supervision. It does not update the model but saves the gradient information. It samples styles like the Discriminator step. See top of Fig.~\ref{fig:full}.
 
    \item \textbf{Autoencoder:} Pairs of images by the same author are concatenated width-wise, and a single style vector is extracted for both of them. Then each image is individually reconstructed using that style. We compute the reconstruction, adversarial and handwriting recognition losses with the reconstructed images. The gradients from this step and the GAN-only step are balanced. Both $S$ and $G$ are updated. See bottom of Fig.~\ref{fig:full}.
\end{enumerate}

We now define the loss functions used in training our model and formalize the gradient balancing.
Let $I$ be a dataset image, $t_I$ its corresponding text, and $c_I$ its corresponding dataset spaced text. 
Let $I'$ be the concatenation of $I$ and another image by the same author.
Let $y_s$ be a sampled style, obtained by sampling two stored style vectors from the running window history and interpolating/extrapolating a point on the line between them.
Let $t_s$ be text sampled from a text corpus.
$MSE$ and $L1$ are mean squared error and L1 loss. $CTC$ is connectionist temporal classification loss \cite{CTC}. $S,G,R,E,C$, and $D$ are the networks described in Sec.~\ref{sec:method}.

\vspace{-0.2in}
\begin{align}
\text{Spacing network loss } l_c &= MSE(C(t_I,S(I')),c_I) \\
\text{Discriminator loss } l_d = max(1-D(I),0) + max(1 &+ D(G(C(t_s,y_s),y_s)),0) \\
\text{Generated image adversarial loss } l_{adv,g} &= -D(G(C(t_s,y_s),y_s)) \\
\text{Generated image recognition loss } l_{rec,g} &=\text{CTC}( R(G(C(t_s,y_s),y_s), t_s ) 
\\
\text{Reconstructed image adversarial loss } l_{adv,r} &= -D(G(c_I,S(I'))) \\
\text{Reconstructed image recognition loss } l_{rec,r} &= \text{CTC}( R(G(c_I,S(I')), t_I)) \\
\text{Combined reconstruction loss } l_{auto,r} = L1(G(c_I,S(I')),I) &+ L1(E(G(c_I,S(I'))),E(I))
\end{align}

$\nabla l_c$ is used to updated $C$ and $S$, and $\nabla l_d$ is used to update $D$. The remaining gradients are balanced. 
Let $m^i_{\nabla l_x}$ be the mean absolute gradient of loss $l_x$ for layer $i$ in the model. 
The gradient of each loss $l_x \in \{l_{adv,g},l_{rec,g},l_{adv,r},l_{rec,r} \} $ is normalized by multiplying each layer $i$'s gradient by $ m^i_{\nabla l_{auto,r}}/m^i_{\nabla l_x}$.
After normalization, the weighted sum $\nabla l_{auto,r} + 0.5(\nabla l_{adv,g}) + 0.6( \nabla l_{rec,g}) + 0.4(\nabla l_{adv,r}) + 0.75(\nabla l_{rec,r})$ is used to updated $G$ and $S$. The weights were chosen heuristically to emphasize the parts we found the model struggled with.


We use a batch size of four, being two pairs of images by the same author for Autoencoder and Spacing steps.
We train our model for 175,000 steps of the curriculum. The stopping point was based on subjective evaluation of the validation set.
We use two Adam optimizers in training; one for the discriminator, and one for the rest of the model (except the pretrained $R$ and $E$).
Both optimizers use a learning rate of 0.0002 and betas of~(0.5,0.999).

\section{Experiments}
 We first discuss the data used. Then we compare to prior methods. We then show exploration into our method with an ablation study (Sec.~\ref{sec:ablation}) and an examination of the style space (Sec.~\ref{sec:space}). We finally discuss a user study we performed (Sec.~\ref{sec:human-study}).

We use the IAM handwriting dataset~\cite{IAM} and the RIMES dataset~\cite{RIMES}, which contain segmented images of handwriting words and lines with accompanying transcriptions.
We developed our method exclusively with the IAM training (6,161 lines) and validation (1,840 lines) sets, and reserved the test sets for experiments (FID/GS scores use training images).
Note that IAM consists of many authors, but authors are disjoint across train/val/test splits.
We resize all images to a fixed height of 64 pixels, maintaining aspect ratio.
We apply a random affine slant transformation to each image in training (-45\textdegree, 45\textdegree ~uniform).

\begin{figure}[t]
\centering
\includegraphics[width=0.78\textwidth]{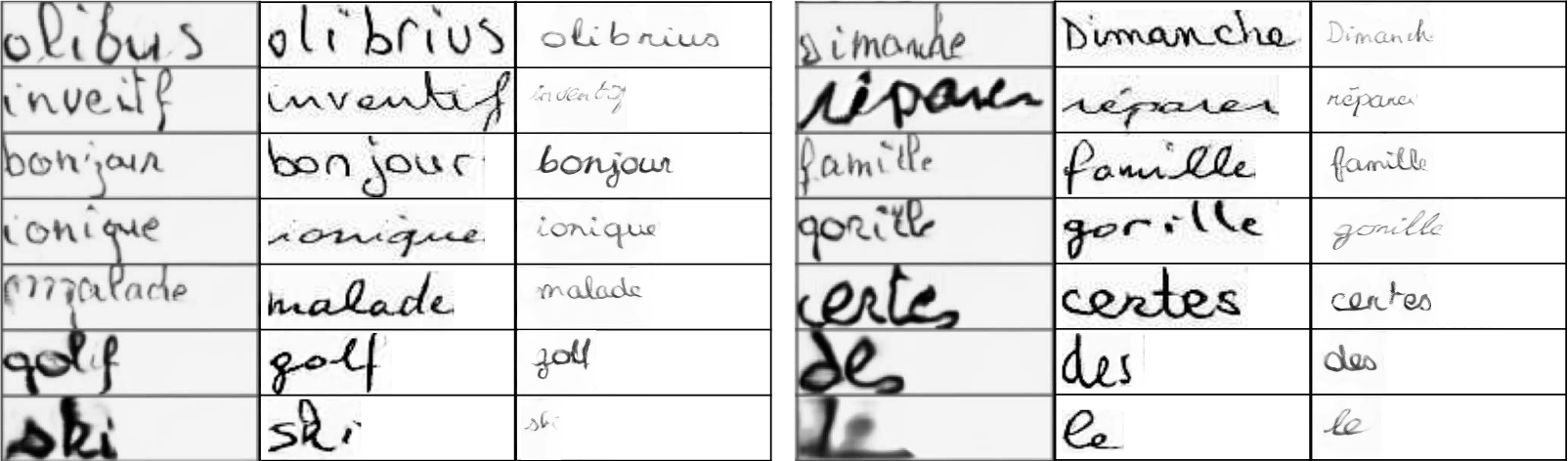}
\vspace{-0.1in}
\caption{Comparing to prior methods on the RIMES dataset. Left: Alonso et al.~\cite{AlonsoGeneration}, middle: ScrabbleGAN \cite{scrabblegan}, right: ours. Our model was trained using full lines, whereas the other two used word images. Segmentation differences caused our model to produce smaller text.
}
\label{fig:words_compare}
\end{figure}

\begin{figure}[t]
\centering
\vspace{-0.05in}
\includegraphics[width=0.65\textwidth]{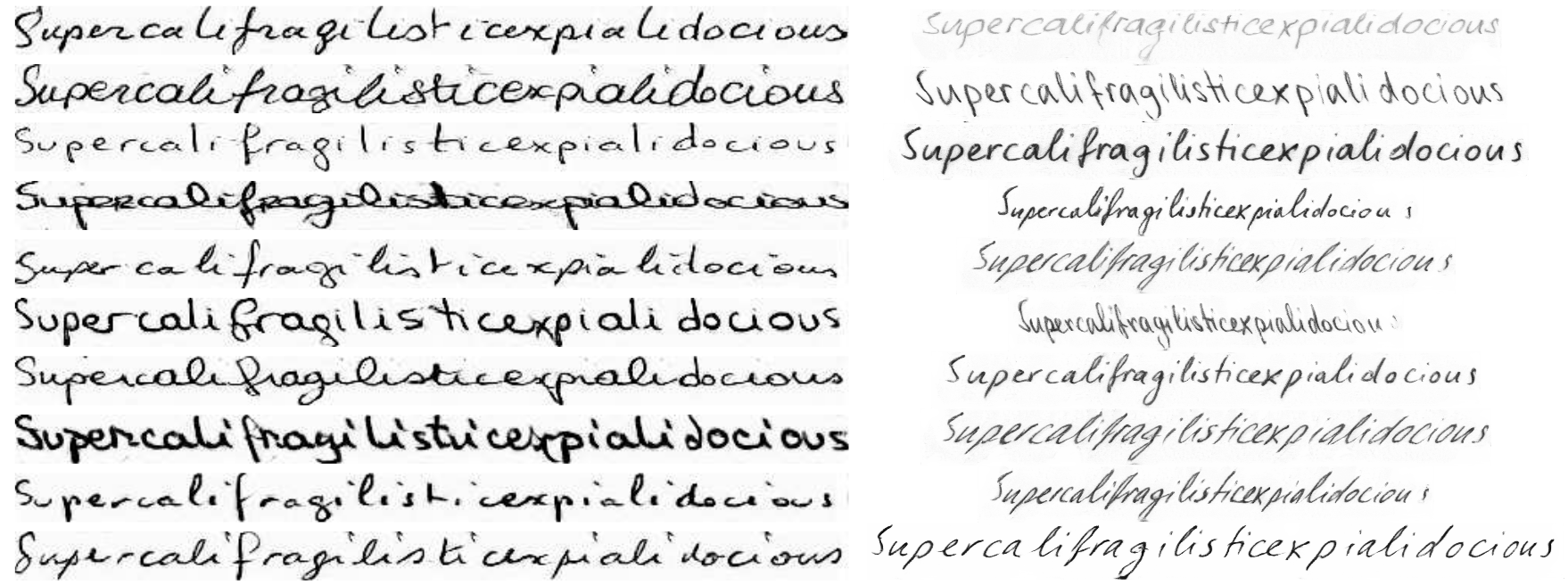}
\vspace{-0.1in}
\caption{Contrasting variability of image length for ScrabbleGAN \cite{scrabblegan} (left) and our method (right) using a fixed word. ScrabbleGAN's horizontal spacing is mostly style agnostic, whereas the spacing in our model is style sensitive.
}
\label{fig:super_compare}
\end{figure}

Fig.~\ref{fig:words_compare} compares our results to those from Alonso et al.~\cite{AlonsoGeneration} and ScrabbleGAN~\cite{scrabblegan}. 
Our results appear to have similar quality as \cite{scrabblegan}. 
It can be seen in Fig.~\ref{fig:super_compare} that \cite{scrabblegan} (left) lacks diversity in horizontal spacing; despite the style changing, the images are always the same length. This is due to their architectural choice to have the length dependant on content, not style. Our method takes both content and style into consideration for spacing, leading to variable length images for the same text.
We report Fr\'{e}chet Inception Distance (FID)~\cite{FID} and Geometry Score (GS)~\cite{GS} in Table~\ref{table:fid} using a setup similar to  
\cite{scrabblegan}. 
There exist some intricacies to the FID and GS calculation which are included in the Supplementary Materials (\ref{sec:fid}). 

\begin{table}[]
\centering
{\small
\begin{tabular}{lllll}
                                                    & Dataset  & FID  & GS \\ \hline
Alonso et al.~\cite{AlonsoGeneration} & RIMES words & 23.94  & $8.58\times 10^{-4}$ \\
ScrabbleGAN \cite{scrabblegan} & RIMES words             & 23.78  & $7.60\times 10^{-4}$ \\
Ours (trained on RIMES lines) & RIMES words                                      & 37.60    & $1.01\times 10^{-1}$ \\ 
Ours (trained on RIMES lines) & RIMES lines                                      & 23.72  & $7.19\times 10^{-1}$ \\ 
Ours (trained on IAM lines) & IAM lines                                      & 20.65    & $4.88\times 10^{-2}$ \\ 
\end{tabular}
}
\caption{FID and GS scores in comparison to prior methods.}
\label{table:fid}
\end{table}

\begin{figure}[t]
    \centering
    \begin{minipage}{0.3\textwidth}
        \centering
        \includegraphics[width=0.72\textwidth]{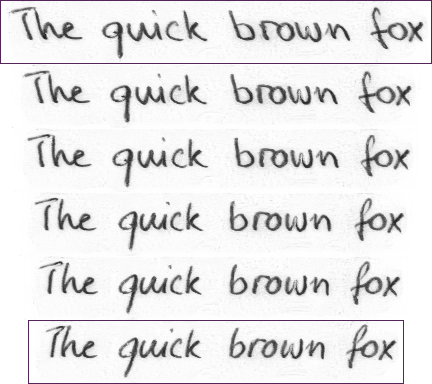} 
        
    \end{minipage}\hfill
    \begin{minipage}{0.3\textwidth}
        \centering
        \includegraphics[width=0.85\textwidth]{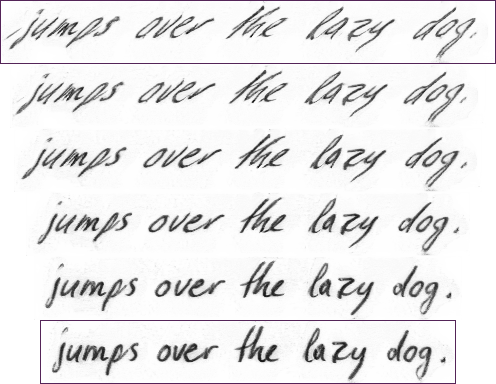} 
    \end{minipage}\hfill
    \begin{minipage}{0.3\textwidth}
        \centering
        \includegraphics[width=0.70\textwidth]{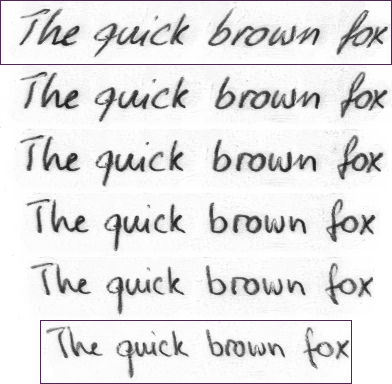} 
    \end{minipage}\hfill
    \caption{Three sets of interpolations between different styles.}
    \label{fig:interpolation}
\end{figure}

\begin{figure}[t]
\centering
\includegraphics[width=0.90\textwidth]{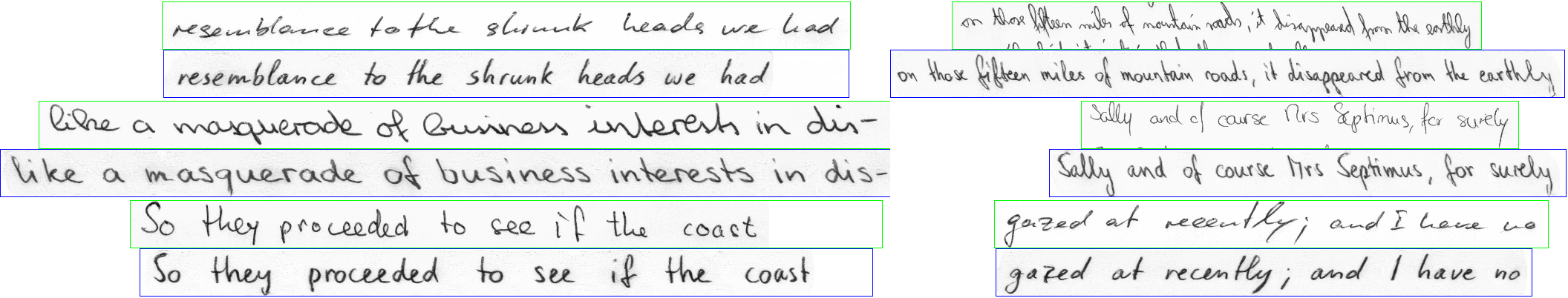}
\vspace{-10pt}
\caption{Reconstruction results. Green is original, blue is our model's reconstruction.
}
\label{fig:recon}
\end{figure}

Fig.~\ref{fig:interpolation} shows interpolation between three sets of two styles taken from test set images. 
These images look realistic, even on the interpolated styles. 
Notice the model even predicts faint background textures similar to dataset images.
We note that while styles vary, it mostly varies in terms of global style elements (e.g., slant, ink thickness); the variation rarely comes from character shapes. 
Figs.~\ref{fig:interpolation} and \ref{fig:super_compare} were generated with text not present in the training set; We notice no difference when generating with text from the dataset compared to other text.
Fig.~\ref{fig:recon} shows reconstruction results of our model. 
The model mimics aspects of global style, but often fails to copy character shape styles (e.g., whether the author loops the letter `l'). 
Additional results are provided as Supplementary Material (\ref{sec:addGen}).

\subsection{Ablation Study}\label{sec:ablation}

\begin{figure}[t!]
\centering
\includegraphics[width=0.85\textwidth]{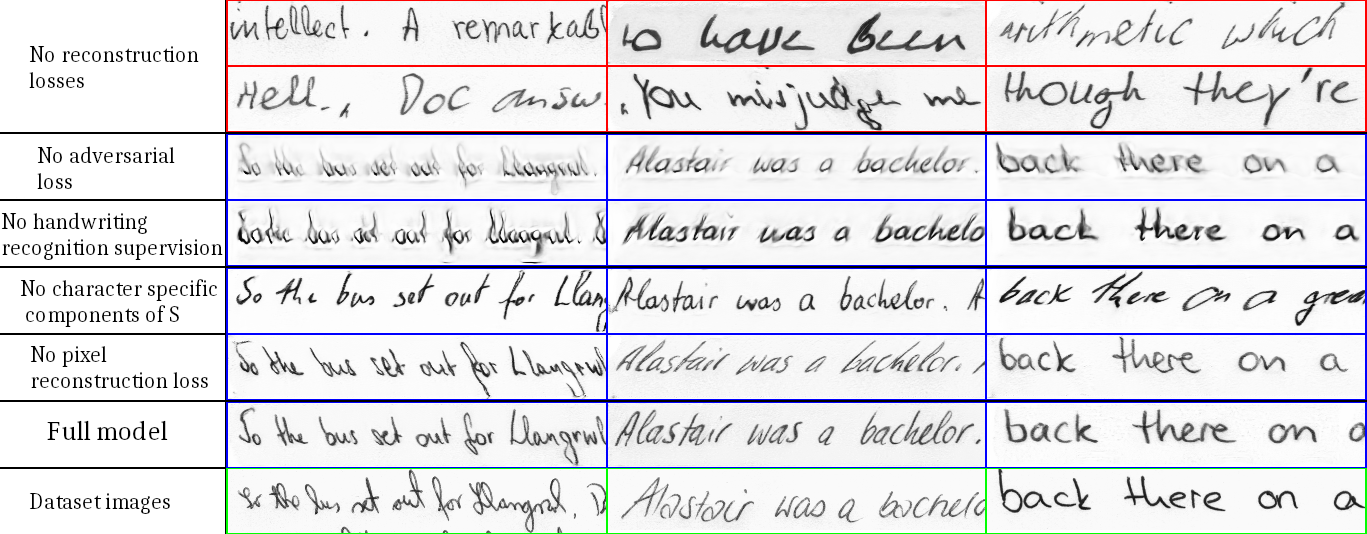}
\vspace{-10pt}
\caption{Ablation study. 
Red images generated using randomly sampled styles. Blue images are attempting to reconstruct the bottom (green) images.  
}
\label{fig:ablation}
\end{figure}

We conducted an ablation study (see Fig.~\ref{fig:ablation}) by removing several key components of our model: the adversarial loss, the handwriting recognition loss, the autoencoding reconstruction losses, and the character specific heads in the style extractor. 
Without the reconstruction loss, the model still generates plausible images and has variety. However, the character shapes are not as well formed. 
Without the adversarial loss, the model produces blurry results.
Curiously, the model produces legible images without the handwriting recognition loss, but with decreased realism. 
The reconstruction loss is likely responsible for legibility, but we are unsure why realism would suffer. 
Without the character specific components of $S$ 
the model loses some ability to mimic styles. 
The pixel reconstruction loss only slightly improves image quality, and without the perceptual loss, the model was unable to converge.

We also were unable produce good results without using spaced text. 
We attempted a model without reconstruction and used 1D convolutions to allow the model to learn the spacing on its own. It failed to produce legible results. 
We attempted to train our model with the gradient balancing of~\cite{AlonsoGeneration}, however the model failed to train.
Earlier versions of our model successfully trained with that gradient balancing, but with decreased quality.
Additional ablation results can be seen in the Supplementary Materials (\ref{sec:addAb})

\begin{figure}[t!]
\begin{center}
 \begin{minipage}[b]{\textwidth}
  \begin{minipage}[b]{0.49\textwidth}
    \centering
    \includegraphics[width=0.70\textwidth]{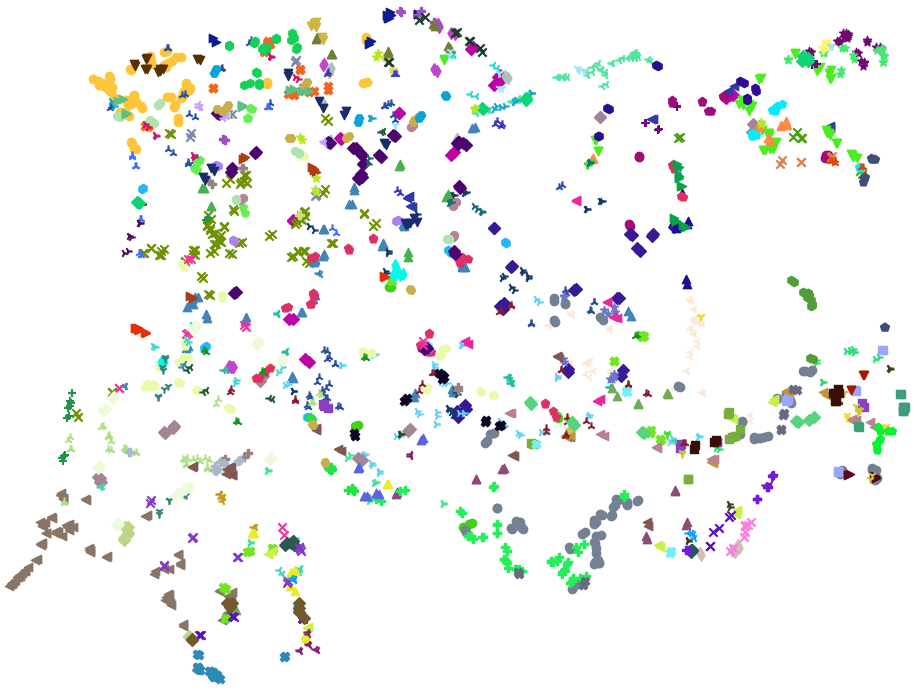}
    \vspace{-7pt}
\captionof{figure}{UMAP projection of the styles extracted from the test set images. Shape and color indicate author. Most styles from the same author cluster together, even though the model was not explicitly trained to do this.
}
\label{fig:umap}
  \end{minipage}
  \hspace{0.02\textwidth}
  \begin{minipage}[b]{0.49\textwidth}
    \centering
    \begin{tabular}{r|ll}
\multicolumn{1}{l|}{} & Guessed: & Guessed: \\
Actually:             & Human    & Computer \\ \cline{2-3} 
Human                 & 34.2\%   & 15.8\%   \\
Generated             & 31.9\%   & 18.0\%   \\ \hline
Poorly                &          &          \\
generated             & 10.5\%   & 89.5\%   
\end{tabular}
\\ ~ \\
\captionof{table}{Top two rows are a confusion matrix of the human study results. Bottom row shows results on deliberately poor generated images as a measure of participant attention.
}
\label{table:conf}
    \end{minipage}
  \end{minipage}
  \end{center}
  \vspace{-0.10in} 
  \end{figure}

\subsection{Latent Style Space}\label{sec:space}

We are able to show evidence that $S$ extracts styles meaningful at the author level.
 In Fig.~\ref{fig:umap} we show a UMAP \cite{UMAP} projection of style vectors extracted from the test set images. Styles extracted from the same author tend to be near each other. There is no specific loss to encourage this behavior; this clustering is learned as the model learns to reconstruct images. Taking the L2 distances between style vectors, the mean distance between styles taken from the same author is 0.916 with a standard deviation of 0.658, and the mean distance between styles taken from different authors is 2.264 with a standard deviation of 1.367.

\subsection{Human Evaluation}
\label{sec:human-study}

We evaluated the realism of our generated images using Amazon Mechanical Turk. 
Participants viewed a single image at a time and were asked if the image was written by a human or a computer. 
Real images were sampled from the test set. 
Generated images used the same text. Styles were interpolated between styles extracted from the test set.
After control measures to ensure participant reliability (described in supplementary material), 
14,875 responses contributed to the final evaluation.
Overall, the participants had an accuracy of 52.2\% at determining whether an image was human or computer generated, indicating that our generated images are generally convincing. 
A confusion matrix of the results is presented in Table~\ref{table:conf}; there is a strong bias towards predicting the images to be human generated. 

We also included deliberately poorly generated images for which we expected close to 100\% accuracy for attentive participants.
Our participants had 89.5\% accuracy on these poorly generated images, indicating that the lack of distinguishability between real and generated images was not simply due to inattention.
While our generated images fooled most participants, we note that the best performing participants (top 10\%) had an average accuracy of 84.9\%, indicating a wide range of participant performance. 
See Supplementary Materials (\ref{sec:human}) for details about this experiment.

\section{Conclusion}

We have presented a system capable of directly generating the pixels of a handwriting image of arbitrary length. 
Our generation is conditioned both on text and style and relies on an spacing network to predict the space needed between text, enabling the generation of arbitrary length images.
Our model is capable of extracting a style from example images and then generating handwriting in that style, but with arbitrary text.
Our method does well at capturing the variations of global style in handwriting, such as slant and size.

\bibliography{egbib}

\begin{thebibliography}{36}
\providecommand{\natexlab}[1]{#1}
\providecommand{\url}[1]{\texttt{#1}}
\expandafter\ifx\csname urlstyle\endcsname\relax
  \providecommand{\doi}[1]{doi: #1}\else
  \providecommand{\doi}{doi: \begingroup \urlstyle{rm}\Url}\fi

\bibitem[Aksan et~al.(2018)Aksan, Pece, and Hilliges]{Aksan:2018:DeepWriting}
Emre Aksan, Fabrizio Pece, and Otmar Hilliges.
\newblock {DeepWriting: Making Digital Ink Editable via Deep Generative
  Modeling}.
\newblock In \emph{SIGCHI Conference on Human Factors in Computing Systems},
  CHI '18, 2018.

\bibitem[Alonso et~al.(2019)Alonso, Moysset, and Messina]{AlonsoGeneration}
Eloi Alonso, Bastien Moysset, and Ronaldo Messina.
\newblock Adversarial generation of handwritten text images conditioned on
  sequences.
\newblock In \emph{15th IAPR International Conference on Document Analysis and
  Recognition (ICDAR)}, Sep 2019.

\bibitem[Chen and Koltun(2017)]{Chen_2017_ICCV}
Qifeng Chen and Vladlen Koltun.
\newblock Photographic image synthesis with cascaded refinement networks.
\newblock In \emph{The IEEE International Conference on Computer Vision
  (ICCV)}, Oct 2017.

\bibitem[Chen et~al.(2009)Chen, Cheng, Tan, Shamir, and Hu]{Sketch2Photo}
Tao Chen, Ming-Ming Cheng, Ping Tan, Ariel Shamir, and Shi-Min Hu.
\newblock Sketch2photo: Internet image montage.
\newblock \emph{ACM Trans. Graph.}, 28\penalty0 (5), 2009.

\bibitem[Choi et~al.(2018)Choi, Choi, Kim, Ha, Kim, and Choo]{choi2018stargan}
Yunjey Choi, Minje Choi, Munyoung Kim, Jung-Woo Ha, Sunghun Kim, and Jaegul
  Choo.
\newblock Star{GAN}: Unified generative adversarial networks for multi-domain
  image-to-image translation.
\newblock In \emph{The IEEE Conference on Computer Vision and Pattern
  Recognition (CVPR)}, 2018.

\bibitem[Fogel et~al.(2020)Fogel, Averbuch-Elor, Cohen, Mazor, and
  Litman]{scrabblegan}
Sharon Fogel, Hadar Averbuch-Elor, Sarel Cohen, Shai Mazor, and Roee Litman.
\newblock Scrabble{GAN}: Semi-supervised varying length handwritten text
  generation.
\newblock In \emph{The IEEE Conference on Computer Vision and Pattern
  Recognition (CVPR)}, June 2020.

\bibitem[Goodfellow et~al.(2014)Goodfellow, Pouget-Abadie, Mirza, Xu,
  Warde-Farley, Ozair, Courville, and Bengio]{goodfellow2014generative}
Ian Goodfellow, Jean Pouget-Abadie, Mehdi Mirza, Bing Xu, David Warde-Farley,
  Sherjil Ozair, Aaron Courville, and Yoshua Bengio.
\newblock Generative adversarial nets.
\newblock In \emph{Advances in Neural Information Processing Systems (NIPS)},
  2014.

\bibitem[Graves(2013)]{Graves2013GeneratingSW}
Alex Graves.
\newblock Generating sequences with recurrent neural networks.
\newblock \emph{ArXiv}, abs/1308.0850, 2013.

\bibitem[Graves and Schmidhuber(2009)]{graves2009offline}
Alex Graves and J{\"u}rgen Schmidhuber.
\newblock Offline handwriting recognition with multidimensional recurrent
  neural networks.
\newblock In \emph{Advances in Neural Information Processing Systems (NIPS)},
  2009.

\bibitem[Graves et~al.(2006)Graves, Fern\'{a}ndez, Gomez, and Schmidhuber]{CTC}
Alex Graves, Santiago Fern\'{a}ndez, Faustino Gomez, and J\"{u}rgen
  Schmidhuber.
\newblock Connectionist temporal classification: Labelling unsegmented sequence
  data with recurrent neural networks.
\newblock In \emph{Proceedings of the 23rd International Conference on Machine
  Learning (ICML)}, 2006.

\bibitem[{Grosicki} and {Abed}(2009)]{RIMES}
E.~{Grosicki} and H.~E. {Abed}.
\newblock {ICDAR} 2009 handwriting recognition competition.
\newblock In \emph{10th IAPR International Conference on Document Analysis and
  Recognition (ICDAR)}, 2009.

\bibitem[Heusel et~al.(2017)Heusel, Ramsauer, Unterthiner, Nessler, and
  Hochreiter]{FID}
Martin Heusel, Hubert Ramsauer, Thomas Unterthiner, Bernhard Nessler, and Sepp
  Hochreiter.
\newblock Gans trained by a two time-scale update rule converge to a local nash
  equilibrium.
\newblock In \emph{Advances in Neural Information Processing Systems (NIPS)
  30}. 2017.

\bibitem[Huang et~al.(2018)Huang, Liu, Belongie, and
  Kautz]{huang2018multimodal}
Xun Huang, Ming-Yu Liu, Serge Belongie, and Jan Kautz.
\newblock Multimodal unsupervised image-to-image translation.
\newblock In \emph{Proceedings of the European Conference on Computer Vision
  (ECCV)}, 2018.

\bibitem[Isola et~al.(2017)Isola, Zhu, Zhou, and Efros]{Isola_2017_CVPR}
Phillip Isola, Jun-Yan Zhu, Tinghui Zhou, and Alexei~A. Efros.
\newblock Image-to-image translation with conditional adversarial networks.
\newblock In \emph{The IEEE Conference on Computer Vision and Pattern
  Recognition (CVPR)}, July 2017.

\bibitem[Ji and Chen(2019)]{Ji2019GenerativeAN}
B.~Ji and Tianyi Chen.
\newblock Generative adversarial network for handwritten text.
\newblock \emph{ArXiv}, abs/1907.11845, 2019.

\bibitem[Johnson et~al.(2016{\natexlab{a}})Johnson, Alahi, and
  Fei-Fei]{johnson2016perceptual}
Justin Johnson, Alexandre Alahi, and Li~Fei-Fei.
\newblock Perceptual losses for real-time style transfer and super-resolution.
\newblock In \emph{Proceedings of the European Conference on Computer Vision
  (ECCV)}, 2016{\natexlab{a}}.

\bibitem[Johnson et~al.(2016{\natexlab{b}})Johnson, Alahi, and
  Fei-Fei]{perceptual}
Justin Johnson, Alexandre Alahi, and Li~Fei-Fei.
\newblock Perceptual losses for real-time style transfer and super-resolution.
\newblock In Bastian Leibe, Jiri Matas, Nicu Sebe, and Max Welling, editors,
  \emph{Proceedings of the European Conference on Computer Vision (ECCV))},
  2016{\natexlab{b}}.

\bibitem[Johnson et~al.()Johnson, Brostow, Shotton, Kwatra, and
  Cipolla]{photosynthesisJohnson}
Matthew Johnson, G.~J. Brostow, J.~Shotton, V.~Kwatra, and R.~Cipolla.
\newblock {Semantic photosynthesis}.
\newblock In Bernice~E. Rogowitz, Thrasyvoulos~N. Pappas, and Scott~J. Daly,
  editors, \emph{Human Vision and Electronic Imaging XII}, volume 6492.
  International Society for Optics and Photonics.

\bibitem[Kang et~al.(2020)Kang, Riba, Wang, Rusinol, Forn{\'e}s, and
  Villegas]{Kang2020GANwritingCG}
Lei Kang, Pau Riba, Yaxing Wang, Marccal Rusinol, Alicia Forn{\'e}s, and
  Mauricio Villegas.
\newblock {GAN}writing: Content-conditioned generation of styled handwritten
  word images.
\newblock In \emph{Proceedings of the European Conference on Computer Vision
  (ECCV)}, August 2020.

\bibitem[Karras et~al.(2019)Karras, Laine, and Aila]{stylegan}
Tero Karras, Samuli Laine, and Timo Aila.
\newblock A style-based generator architecture for generative adversarial
  networks.
\newblock In \emph{The IEEE Conference on Computer Vision and Pattern
  Recognition (CVPR)}, June 2019.

\bibitem[Karras et~al.(2020)Karras, Laine, Aittala, Hellsten, Lehtinen, and
  Aila]{karras2019analyzing}
Tero Karras, Samuli Laine, Miika Aittala, Janne Hellsten, Jaakko Lehtinen, and
  Timo Aila.
\newblock Analyzing and improving the image quality of style{GAN}.
\newblock In \emph{The IEEE Conference on Computer Vision and Pattern
  Recognition (CVPR)}, 2020.

\bibitem[Khrulkov and Oseledets(2018)]{GS}
Valentin Khrulkov and Ivan Oseledets.
\newblock Geometry score: A method for comparing generative adversarial
  networks.
\newblock In \emph{Proceedings of the 35th International Conference on Machine
  Learning (PMLR)}, 2018.

\bibitem[Larsen et~al.(2016)Larsen, Sønderby, Larochelle, and
  Winther]{vae-auto-larsen16}
Anders Boesen~Lindbo Larsen, Søren~Kaae Sønderby, Hugo Larochelle, and Ole
  Winther.
\newblock Autoencoding beyond pixels using a learned similarity metric.
\newblock In \emph{Proceedings of the 33rd International Conference on Machine
  Learning (PMLR)}, Jun 2016.

\bibitem[Liu et~al.(2019)Liu, Huang, Mallya, Karras, Aila, Lehtinen, and
  Kautz]{funit}
Ming-Yu Liu, Xun Huang, Arun Mallya, Tero Karras, Timo Aila, Jaakko Lehtinen,
  and Jan Kautz.
\newblock Few-shot unsupervised image-to-image translation.
\newblock In \emph{Proceedings of the IEEE/CVF International Conference on
  Computer Vision (ICCV)}, October 2019.

\bibitem[Marti and Bunke(2002)]{IAM}
U-V Marti and Horst Bunke.
\newblock The {IAM}-database: an {English} sentence database for offline
  handwriting recognition.
\newblock \emph{International Journal on Document Analysis and Recognition},
  5\penalty0 (1), 2002.

\bibitem[McInnes et~al.(2018)McInnes, Healy, and Melville]{UMAP}
Leland McInnes, John Healy, and James Melville.
\newblock Umap: Uniform manifold approximation and projection for dimension
  reduction.
\newblock \emph{arXiv preprint arXiv:1802.03426}, 2018.

\bibitem[Mellor et~al.(2019)Mellor, Park, Ganin, Babuschkin, Kulkarni,
  Rosenbaum, Ballard, Weber, Vinyals, and Eslami]{Mellor2019UnsupervisedDA}
John F.~J. Mellor, Eunbyung Park, Yaroslav Ganin, Igor Babuschkin, Tejas
  Kulkarni, Dan Rosenbaum, Andy Ballard, Th{\'e}ophane Weber, Oriol Vinyals,
  and S.~M.~Ali Eslami.
\newblock Unsupervised doodling and painting with improved spiralq.
\newblock \emph{ArXiv}, abs/1910.01007, 2019.

\bibitem[Mirza and Osindero(2014)]{mirza2014conditional}
Mehdi Mirza and Simon Osindero.
\newblock Conditional generative adversarial nets.
\newblock \emph{arXiv preprint arXiv:1411.1784}, 2014.

\bibitem[Nakano(2019)]{Nakano2019NeuralPA}
Reiichiro Nakano.
\newblock Neural painters: A learned differentiable constraint for generating
  brushstroke paintings.
\newblock In \emph{Machine Learning for Creativity and Design (NeurIPS
  workshop}, Dec 2019.

\bibitem[Park et~al.(2019)Park, Liu, Wang, and Zhu]{guagan}
Taesung Park, Ming-Yu Liu, Ting-Chun Wang, and Jun-Yan Zhu.
\newblock Semantic image synthesis with spatially-adaptive normalization.
\newblock In \emph{The IEEE Conference on Computer Vision and Pattern
  Recognition (CVPR)}, June 2019.

\bibitem[Puigcerver(2017)]{Puigcerver}
J.~Puigcerver.
\newblock Are multidimensional recurrent layers really necessary for
  handwritten text recognition?
\newblock In \emph{14th IAPR International Conference on Document Analysis and
  Recognition ({ICDAR})}, Nov 2017.

\bibitem[Wang et~al.(2018)Wang, Liu, Zhu, Tao, Kautz, and
  Catanzaro]{Wang_2018_CVPR}
Ting-Chun Wang, Ming-Yu Liu, Jun-Yan Zhu, Andrew Tao, Jan Kautz, and Bryan
  Catanzaro.
\newblock High-resolution image synthesis and semantic manipulation with
  conditional gans.
\newblock In \emph{The IEEE Conference on Computer Vision and Pattern
  Recognition (CVPR)}, June 2018.

\bibitem[{Wigington} et~al.(2017){Wigington}, {Stewart}, {Davis}, {Barrett},
  {Price}, and {Cohen}]{augmentation}
C.~{Wigington}, S.~{Stewart}, B.~{Davis}, B.~{Barrett}, B.~{Price}, and
  S.~{Cohen}.
\newblock Data augmentation for recognition of handwritten words and lines
  using a cnn-lstm network.
\newblock In \emph{14th IAPR International Conference on Document Analysis and
  Recognition (ICDAR)}, 2017.

\bibitem[Wigington et~al.(2018)Wigington, Tensmeyer, Davis, Barrett, Price, and
  Cohen]{startfollowread}
Curtis Wigington, Chris Tensmeyer, Brian Davis, William Barrett, Brian Price,
  and Scott Cohen.
\newblock Start, follow, read: End-to-end full-page handwriting recognition.
\newblock In \emph{Proceedings of the European Conference on Computer Vision
  (ECCV)}, September 2018.

\bibitem[Zhang et~al.(2017)Zhang, Xu, Li, Zhang, Wang, Huang, and
  Metaxas]{Zhang_2017_ICCV}
Han Zhang, Tao Xu, Hongsheng Li, Shaoting Zhang, Xiaogang Wang, Xiaolei Huang,
  and Dimitris~N. Metaxas.
\newblock Stackgan: Text to photo-realistic image synthesis with stacked
  generative adversarial networks.
\newblock In \emph{The IEEE International Conference on Computer Vision
  (ICCV)}, Oct 2017.

\bibitem[Zhang et~al.(2018)Zhang, Xu, Li, Zhang, Wang, Huang, and
  Metaxas]{zhang2018stackgan++}
Han Zhang, Tao Xu, Hongsheng Li, Shaoting Zhang, Xiaogang Wang, Xiaolei Huang,
  and Dimitris~N Metaxas.
\newblock Stackgan++: Realistic image synthesis with stacked generative
  adversarial networks.
\newblock \emph{IEEE Transactions on Pattern Analysis and Machine Intelligence
  (PAMI)}, 41\penalty0 (8), 2018.

\end{thebibliography}


\newpage


\section*{Supplementary Material}
\beginsupplement

\renewcommand\thesubsection{S.\arabic{subsection}}

This document provides supplementary material for the paper ``Text and Style Conditioned GAN for Generation of Offline-Handwriting Lines'' submitted to BMVC 2020, including details of the human study described in the paper, additional image results, additional experimental ablation study results, and architectural details for the networks described in the paper. The sections are as follows:

\begin{itemize}

    \item \ref{sec:fid} Details on FID (and GS) computation.
    \item \ref{sec:human} Details on human experiment.
    \item \ref{sec:addGen} Additional generation results.
    \item \ref{sec:addAb} Additional ablation results.
    \item \ref{sec:model_supp} Network specifications of each model part.
\end{itemize}

\subsection[S.1]{Discussion of FID evaluation and GS details}
\label{sec:fid}

FID \cite{FID} is evaluated by passing an image through the convolutional network Inception-v3 and computing statistics on the average pooled features. Inception-v3 was designed to accept images of size $299 \times 299$, and thus most implementations of FID rescale images to this size before feeding them to the network. In most situations this is fine since GANs typically generate square images. However, in the case of handwriting, particularly lines, images are generally much wider than they are tall. Resizing them to be square causes significant distortions to the image. 
Thus, it would make sense to resize images to a height of 299 and maintain the aspect ratio. Since Inception-v3 is fully convolutional up to the average pooling, it can accept variable sized images. We evaluated FID with both the original square resizing and aspect ratio preserving resizing. We found the scores produced when preserving the aspect ratio appeared closest to the FID reported in \cite{AlonsoGeneration} and \cite{scrabblegan} and thus assume these authors applied something similar, although they do not report this.
We follow \cite{AlonsoGeneration} in using 25,000 training set images and generate 25,000 images using the same lexicon (words or lines depending on dataset), but styles extracted from the test set. Like \cite{scrabblegan}, we only run the experiment once.

When comparing our generated images to RIMES words, there is a  distribution difference caused by segmentation differences. RIMES words are segmented tightly to each word. Our model is trained on RIMES lines, which generally have more whitespace on the top and bottom of each word. Fig.~\ref{fig:words_compare} demonstrates this difference. To make comparison more fair, we crop our generated words on the top and bottom to the first ink pixel (value less than 200). This cropping resembles the segmentation of the word images and improves our FID score.

We also question in general the validity of using FID score for handwriting images. As Inception-v3 is trained on natural images, not handwriting, FID seems ill-suited for evaluating the quality of handwriting images. Further investigation is required into the topic of applying FID to image domains other than natural images.

For GS~\cite{GS}, the data is expected to all be the same size. Because the dataset has variable width images and our method produces variable width images, we pad images to be the same width. Neither \cite{AlonsoGeneration} nor \cite{scrabblegan} report how they handled this. We resize our images to a height of 32 to match \cite{AlonsoGeneration} and \cite{scrabblegan}. Like~\cite{scrabblegan}, we only run the experiment once.


\clearpage

\subsection{Human Study Details}
\label{sec:human}

We submitted 78 image tasks to Amazon Mechanical Turk (35 real, 35 generated, 8 poorly generated), requesting 200 workers to review each image. Each task consisted of instructions, with example images, a task image (real, generated, or poorly generated) and two multiple choice questions. The first question asked the worker to select the correct transcription for the task image. Two choices were shown, one with the correct transcription, the other a permutation of the correct transcription's words (where the first and last words remained in the same place). We removed punctuation so the permutation didn't create artifacts that made the choice too easy. This was to ensure the worker actually looked at the image and was paying attention to what they were doing. The second asked if they thought the image was written by a human or a computer.

\begin{figure}[p]
\centering
\includegraphics[width=0.99\textwidth]{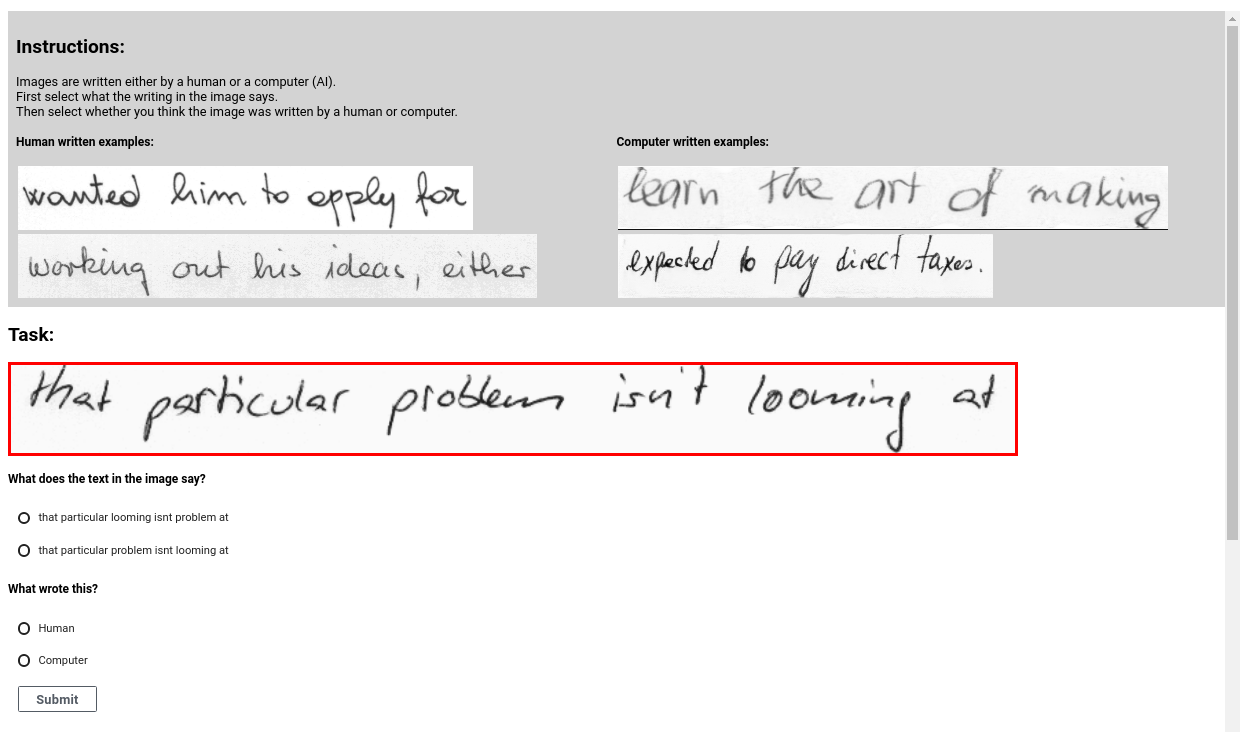}
\caption{A screenshot of the interface the workers saw when completing a task. The example images remained the same each task. The order in which the correct and incorrect transcription responses were placed was random. We kept the task image large so detail could be seen.
}
\label{fig:screenshot}
\end{figure}

The interface the workers saw can be seen in Fig.~\ref{fig:screenshot}. The correct and incorrect transcription options were randomly ordered, the options between human and computer remained in the same order.

The real instances used in the study were randomly selected from the test set. The generated images used the same text as the selected real instances, but the styles were from interpolations between styles extracted from randomly selected test set images. 

\begin{figure}[p]
\centering
\includegraphics[width=0.8\textwidth]{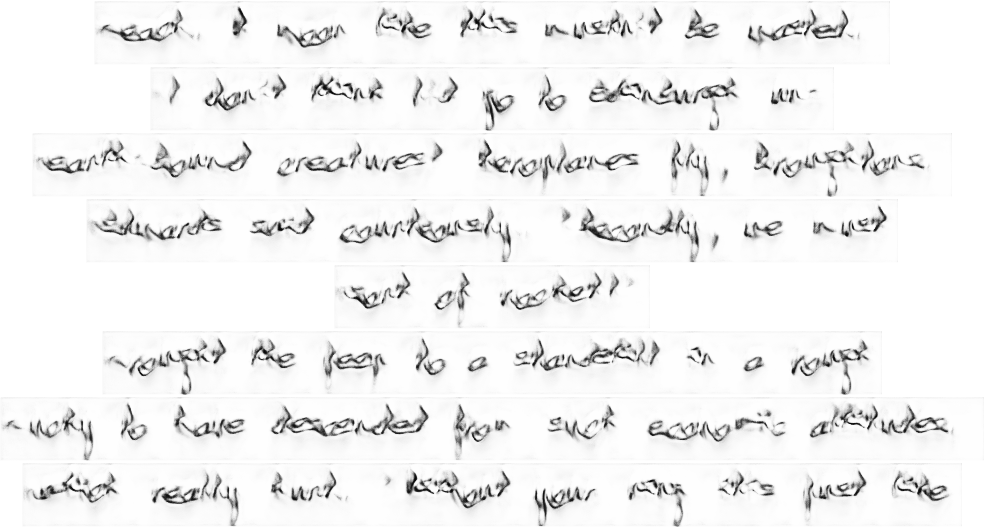}
\caption{Poorly generated images from an intentionally under-trained model used in human study to evaluate participant ability or attention.  These samples are \emph{not} from our final model. }
\label{fig:poor_images}
\end{figure}

\begin{figure}[p]
\centering
\includegraphics[width=0.8\textwidth]{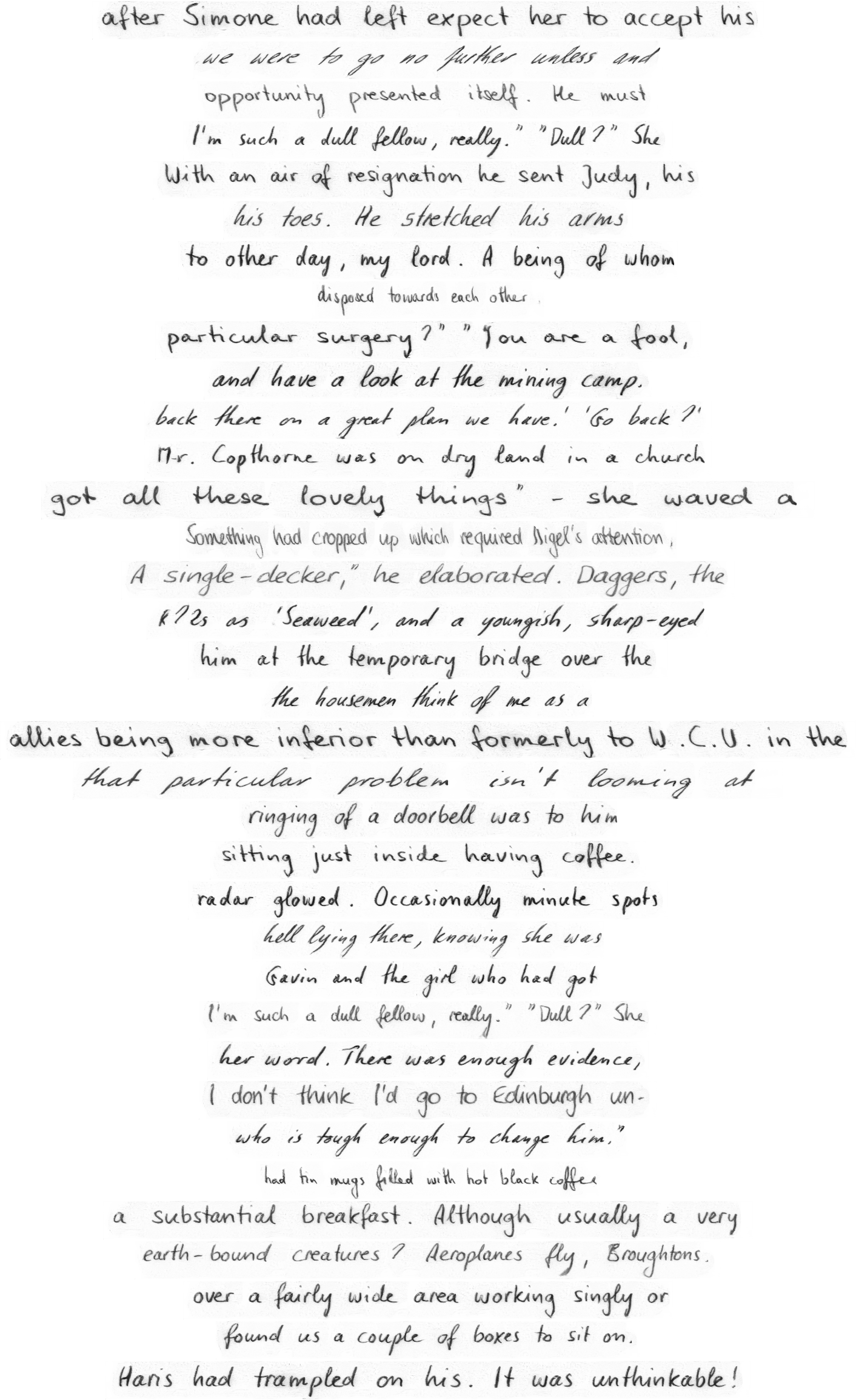}
\caption{Generated images used in human study that were generated using random styles (i.e. random interpolation of style vectors extracted from random pairs of real images from IAM) and random text from the IAM corpus.}
\label{fig:fake_images}
\end{figure}

\begin{figure}[p]
\centering
\includegraphics[width=0.8\textwidth]{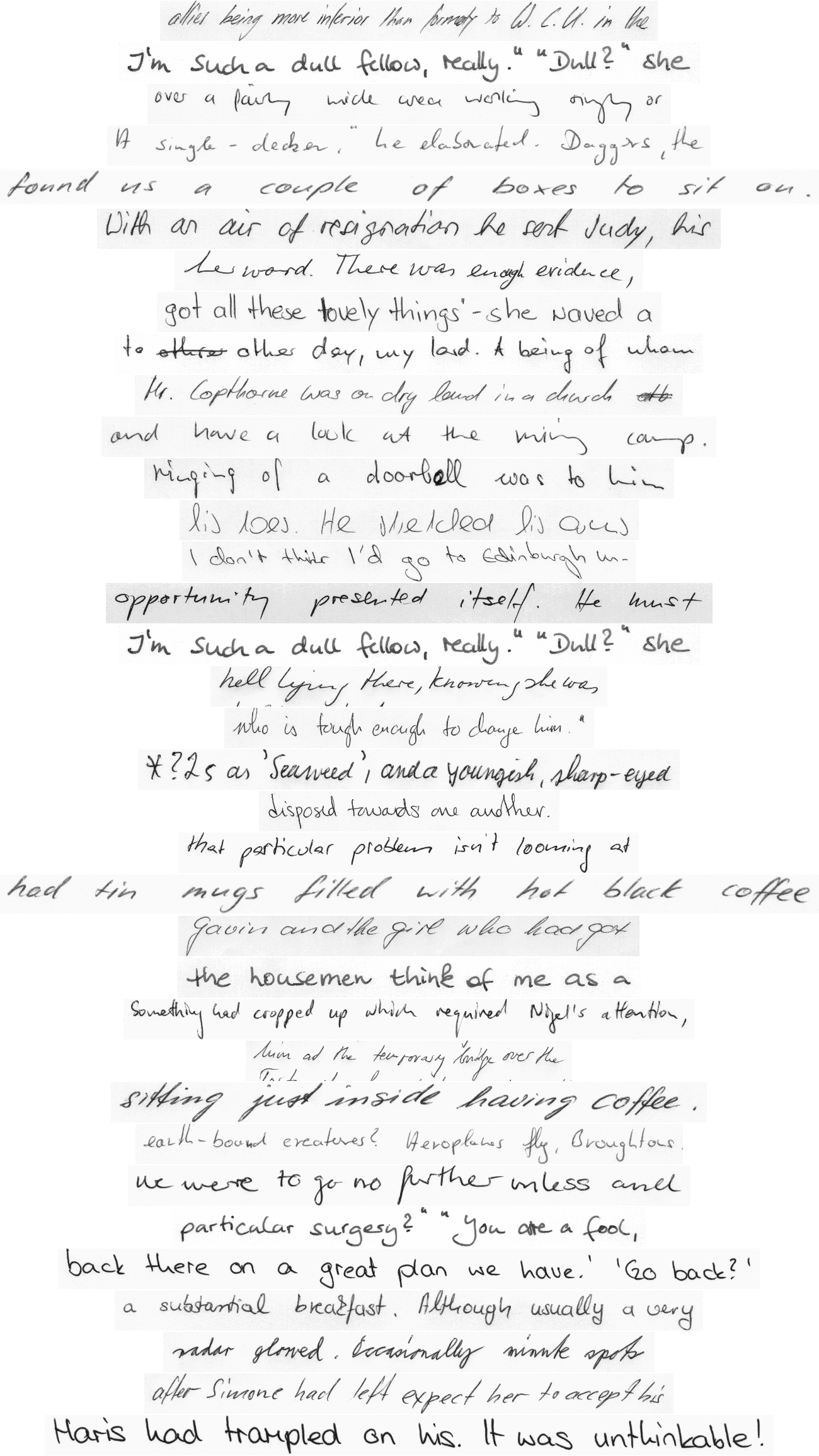}
\caption{Dataset images used in human study.  These are randomly sampled from IAM.}
\label{fig:real_images}
\end{figure}

To help measure the reliability of the workers, we included poorly generated images which should appear to not be written by a human. These were created using a model only trained 2,000 iterations. The responses on these images were not included in the final evaluation, but were held out to help gauge the confidence that can be placed in the workers efforts. The poorly generated images used in the study are shown in Fig.~\ref{fig:poor_images}. The generated and dataset images used in the study are in Figs.~\ref{fig:fake_images} and~\ref{fig:real_images} respectively.

The transcription question was used to filter out workers which were unreliable (likely clicking random responses to complete the tasks quickly). We only used workers who had at least 90\% accuracy on transcription (permutations can sometimes be very close to the correct transcription leading to some error in even engaged workers). Additionally, we only used workers we had at least 6 responses for. The selected workers had 89.5\% accuracy on the poorly generated images, the left-out workers had 79.0\% accuracy.

\clearpage

\subsection{Additional Generation Results}
\label{sec:addGen}

\begin{figure}[p]
\centering
\includegraphics[width=0.65\textwidth]{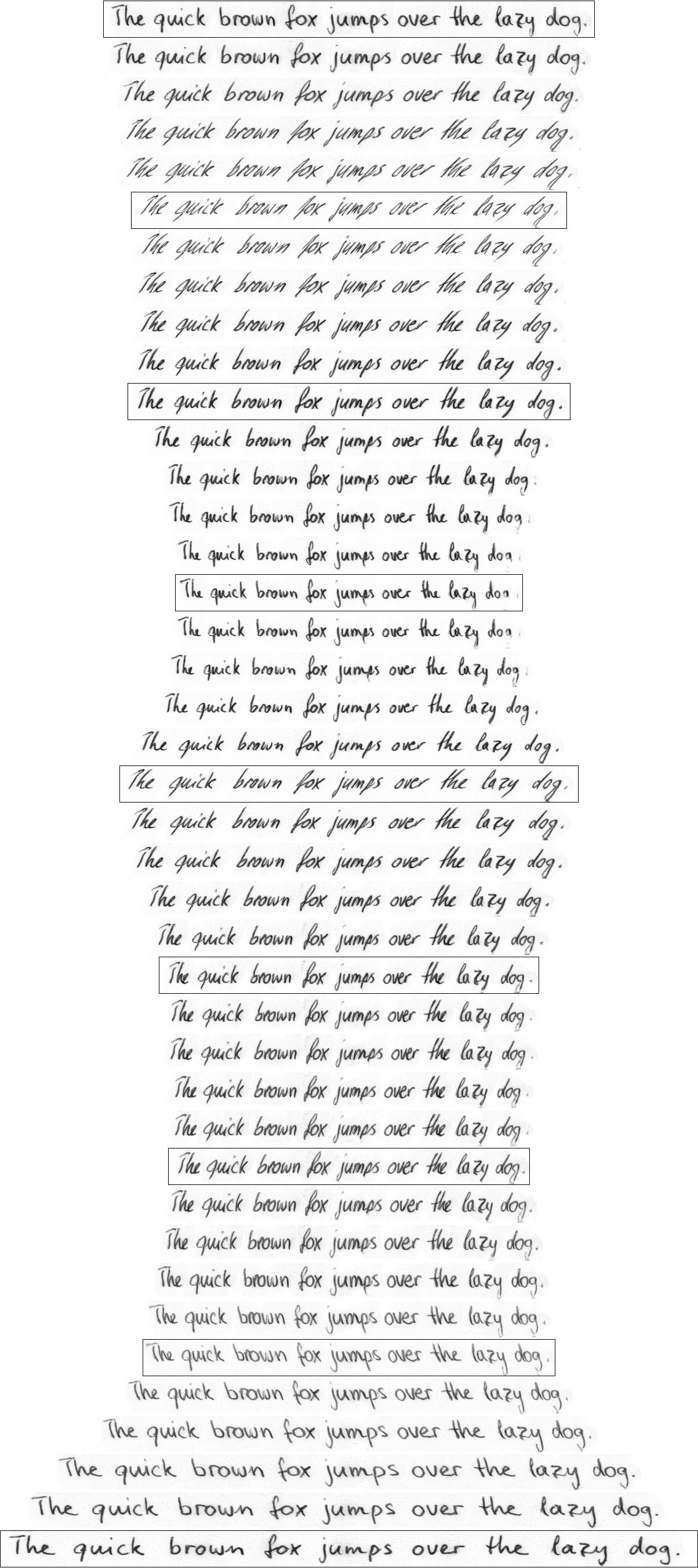}
\caption{Additional interpolation results between 9 different styles extracted from test set images.}
\label{fig:supp_inter}
\end{figure}

\begin{figure}[p]
\centering
\includegraphics[width=0.63\textwidth]{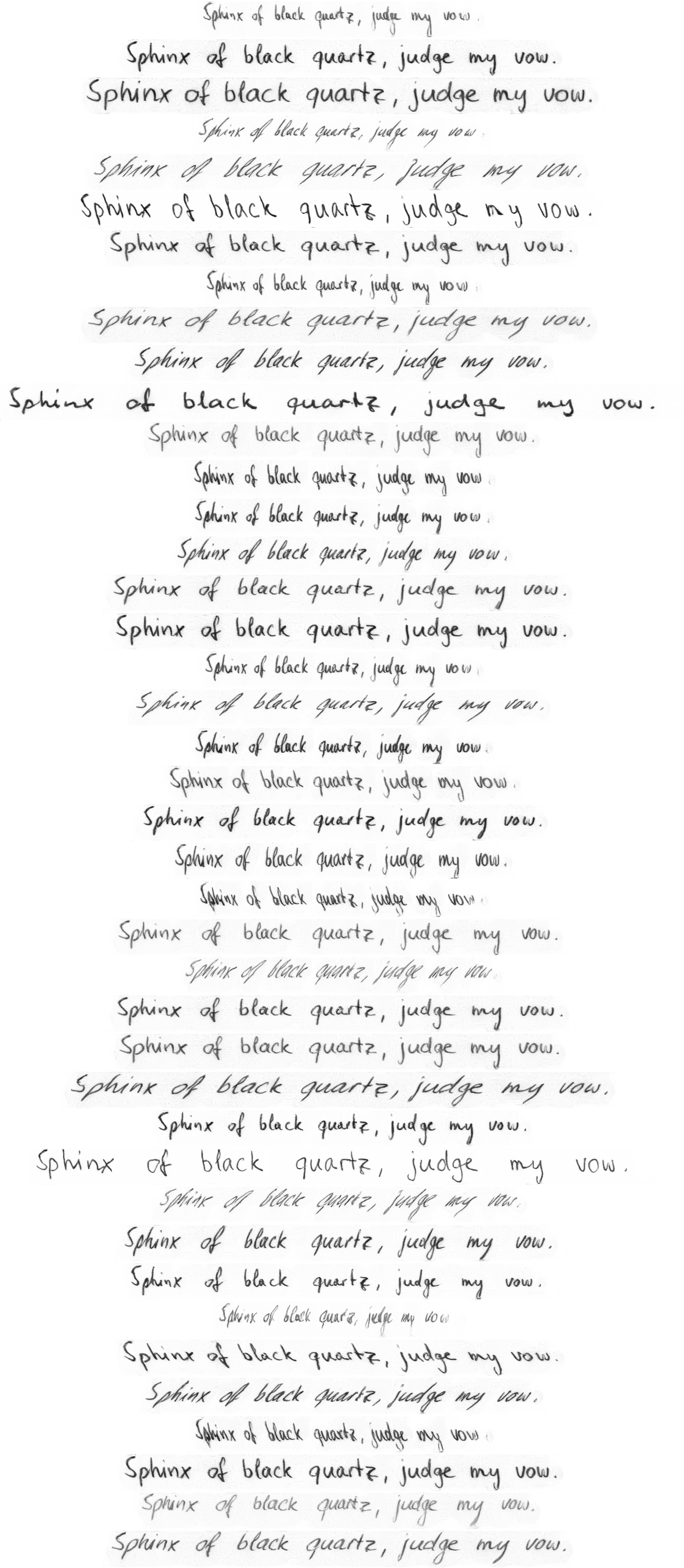}
\caption{Additional generation results using random extra/interpolations between test set styles using the same text.}
\label{fig:supp_random}
\end{figure}

\begin{figure}[p]
\centering
\includegraphics[width=0.85\textwidth]{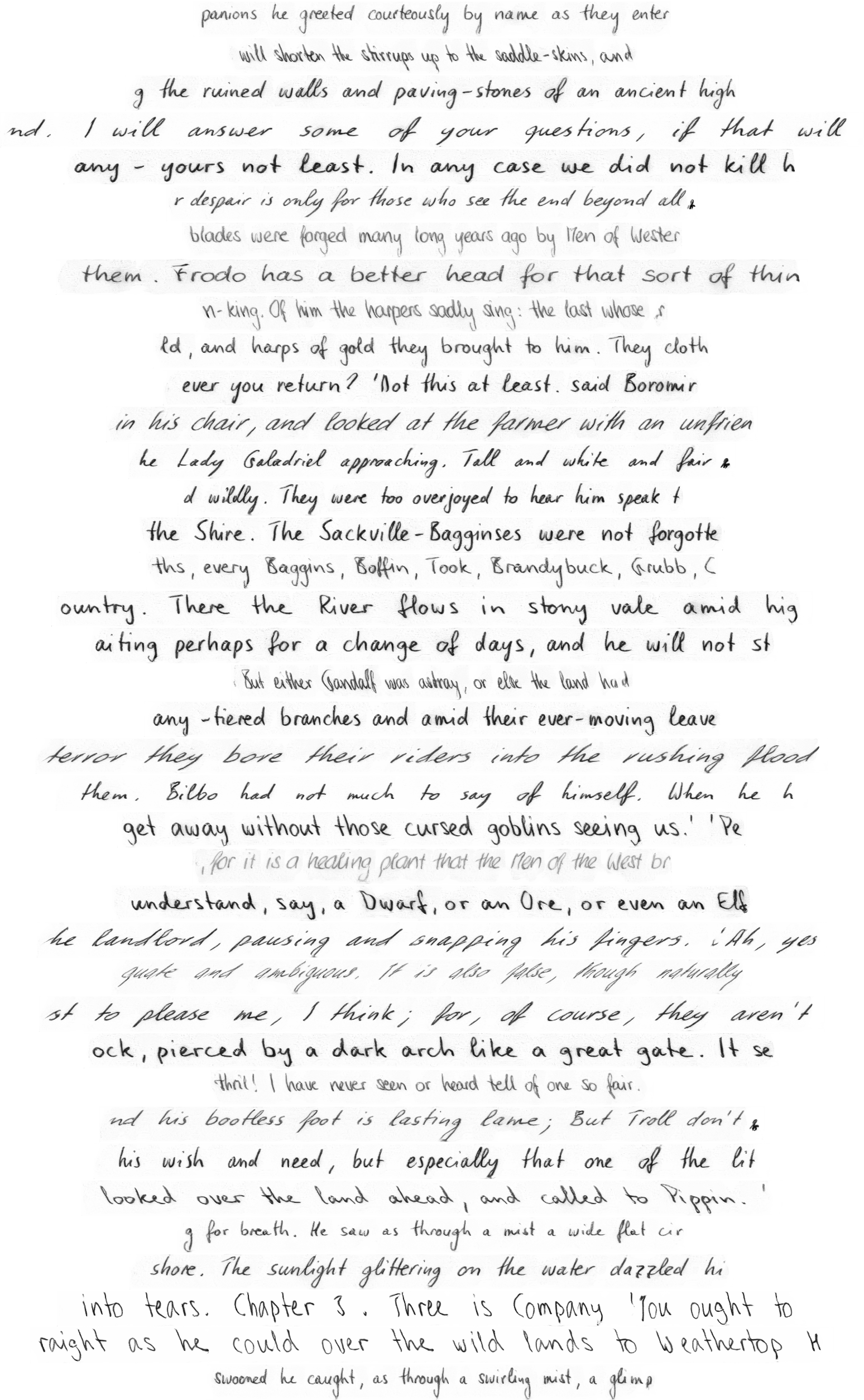}
\caption{Additional generation results using random extra/interpolations between test set styles using varying text.}
\label{fig:supp_random_random}
\end{figure}

\begin{figure}[p]
\centering
\includegraphics[width=0.7\textwidth]{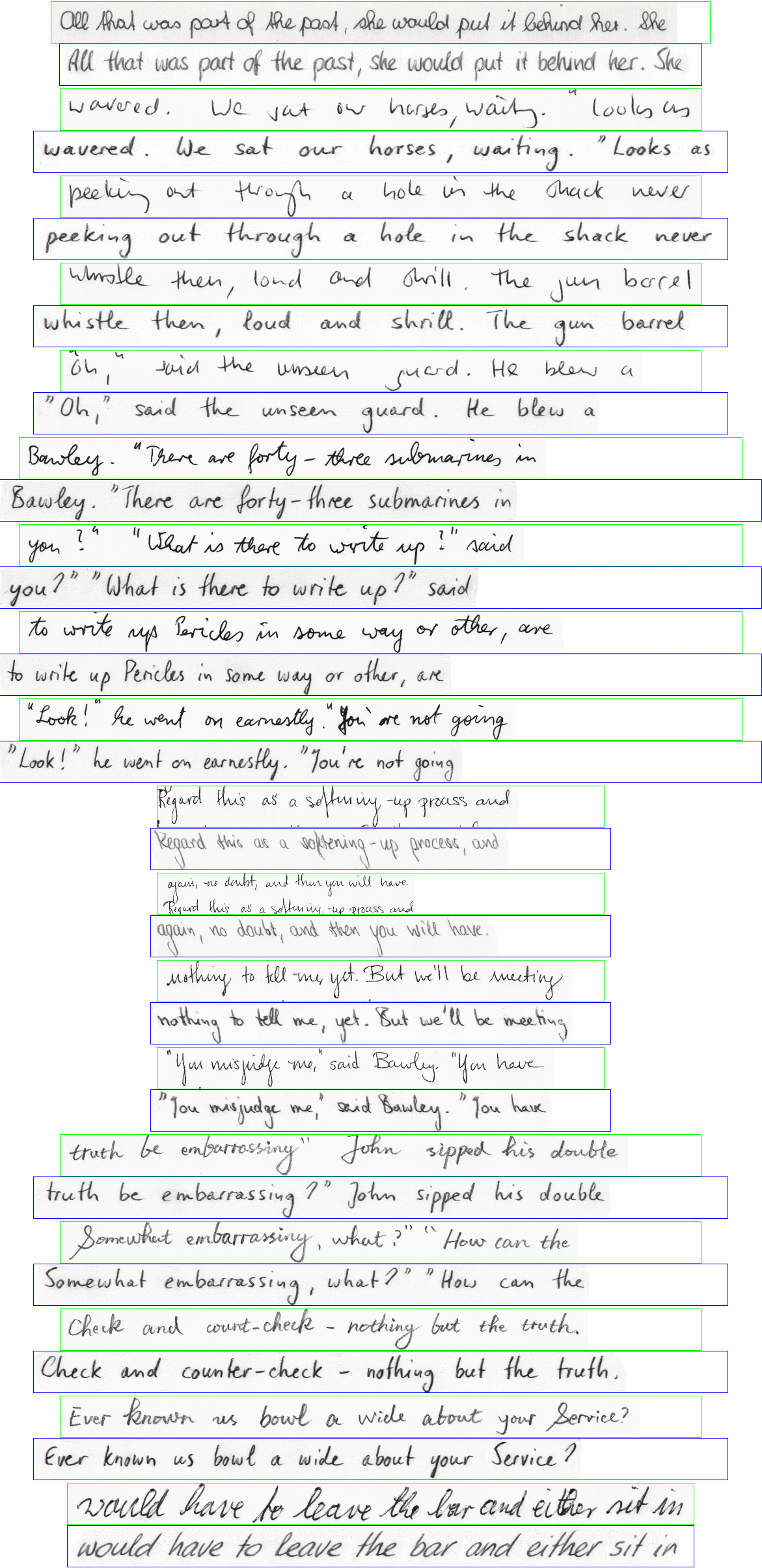}
\caption{Additional Reconstruction results. Green is original, blue is our model's reconstruction.}
\label{fig:supp_recon}
\end{figure}

\begin{figure}[h]
\centering
\includegraphics[width=0.7\textwidth]{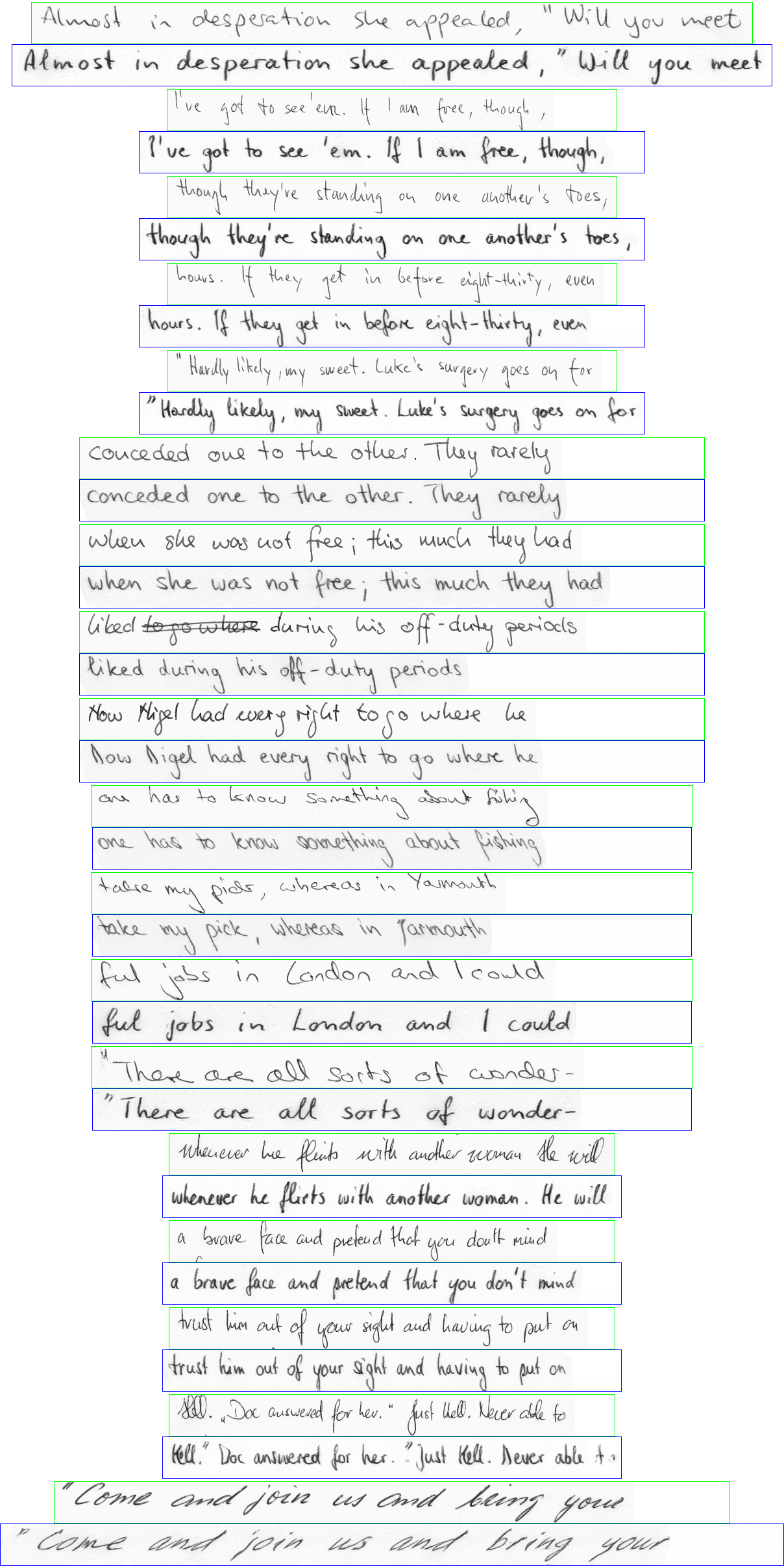}
\caption{Additional Reconstruction results. Green is original, blue is our model's reconstruction.}
\label{fig:supp_recon2}
\end{figure}

We here show additional results from our model. Fig.~\ref{fig:supp_inter} shows additional examples of style interpolation. Figs.~\ref{fig:supp_random} and~\ref{fig:supp_random_random} shows generation using random interpolated/extrapolated styles with fixed and varying text respectively. Figs.~\ref{fig:supp_recon} and~\ref{fig:supp_recon2} show reconstruction results.

        
    

\clearpage

\subsection{Additional Ablation Results}
\label{sec:addAb}

\begin{figure}[h]
\centering
\includegraphics[width=0.8\textwidth]{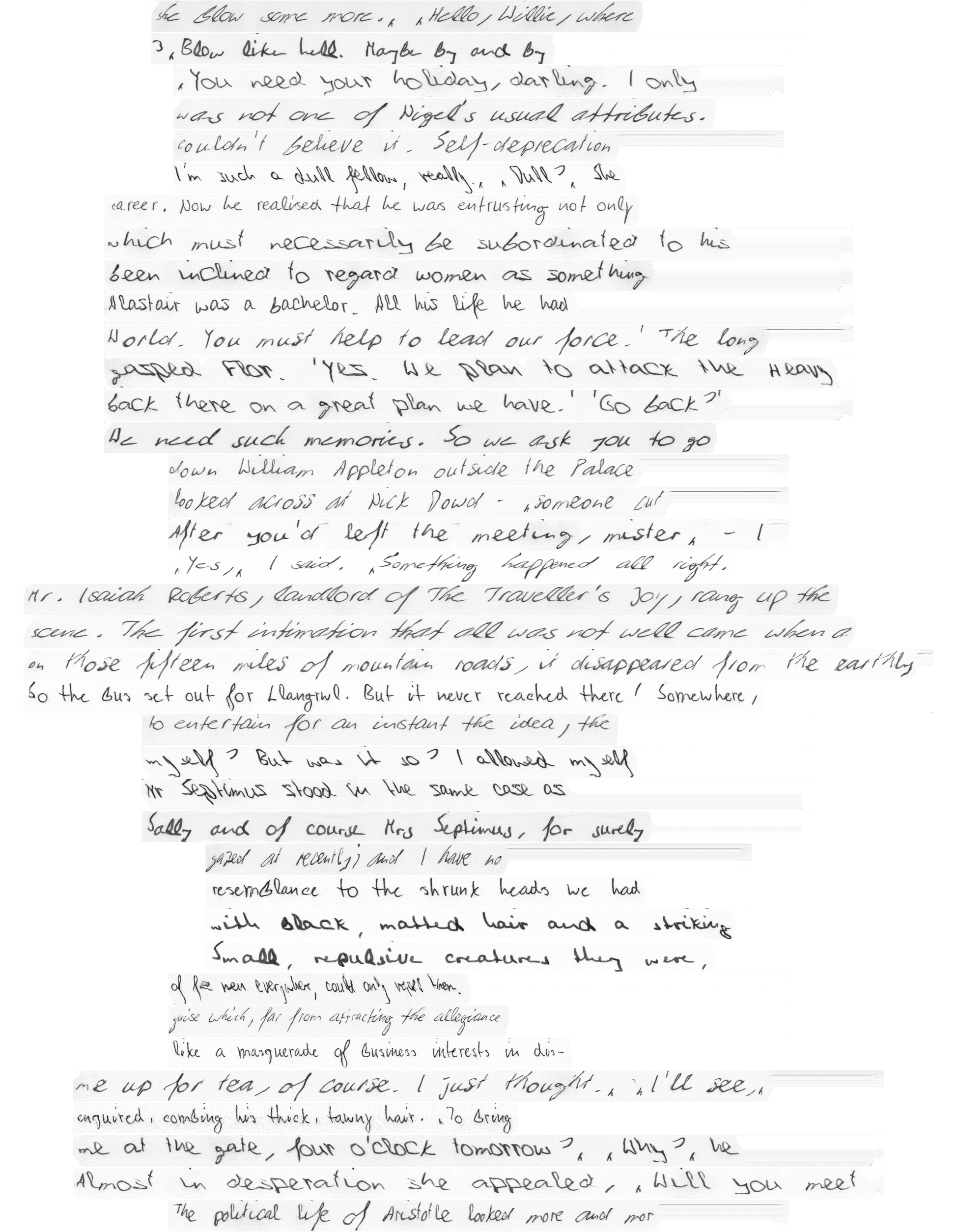}
\caption{Additional ablation results, without the reconstruction losses (random styles).}
\label{fig:supp_noAuto}
\end{figure}

\begin{figure}[h]
\centering
\includegraphics[width=0.7\textwidth]{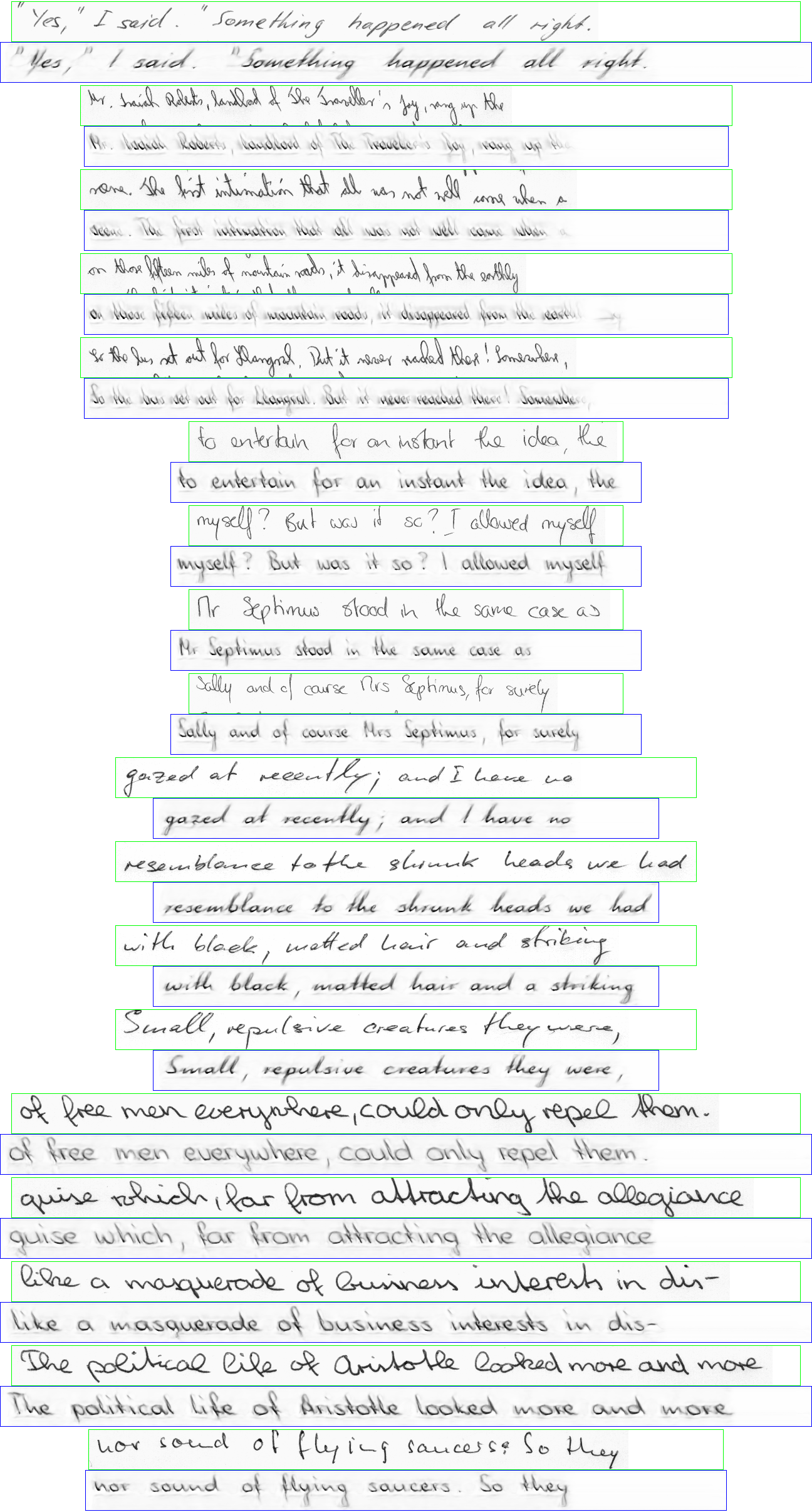}
\caption{Additional ablation results, without adversarial loss.}
\label{fig:supp_noGAN}
\end{figure}

\begin{figure}[h]
\centering
\includegraphics[width=0.7\textwidth]{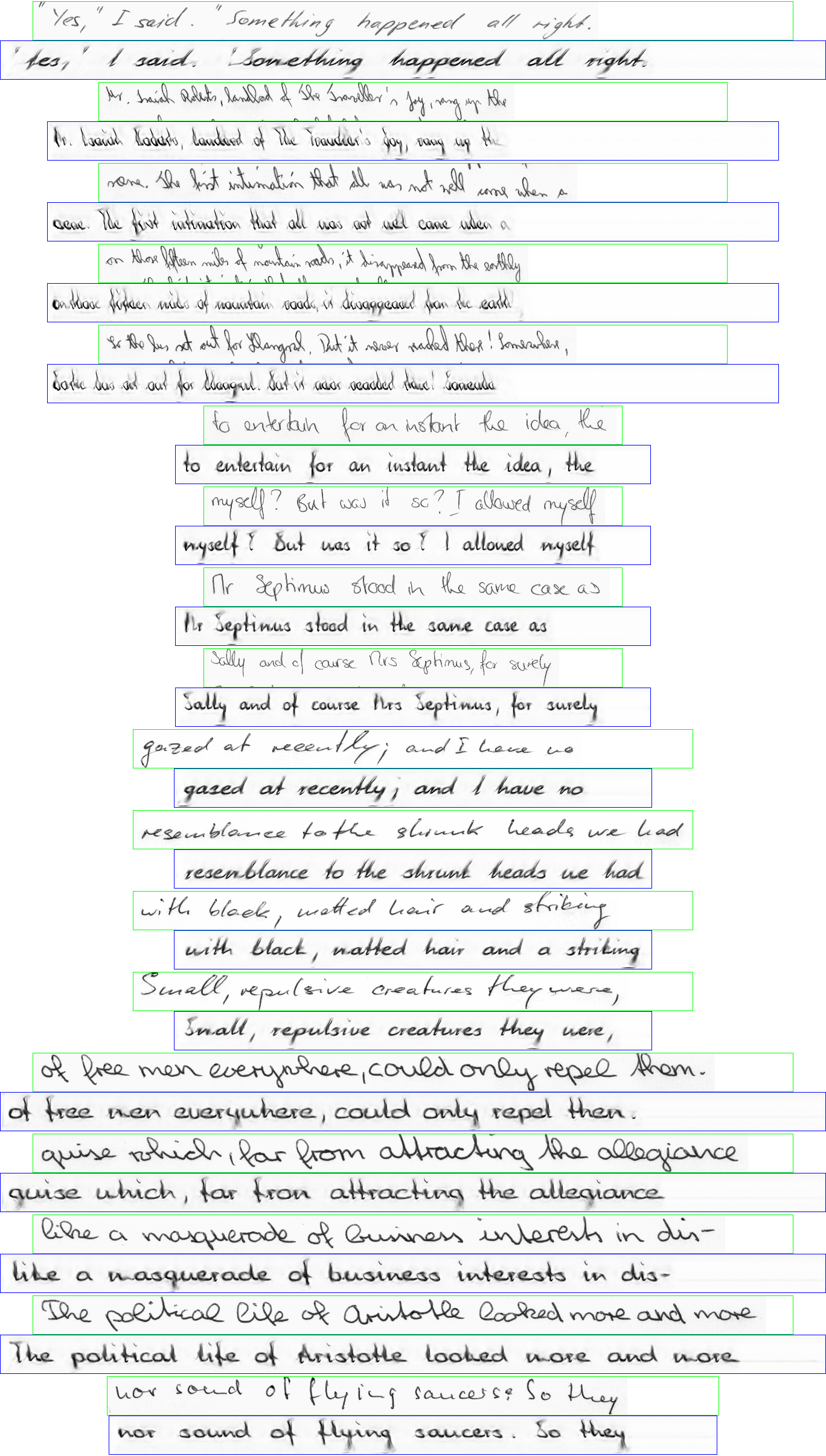}
\caption{Additional ablation results, without handwriting recognition supervision.}
\label{fig:supp_noHWR}
\end{figure}

\begin{figure}[h]
\centering
\includegraphics[width=0.7\textwidth]{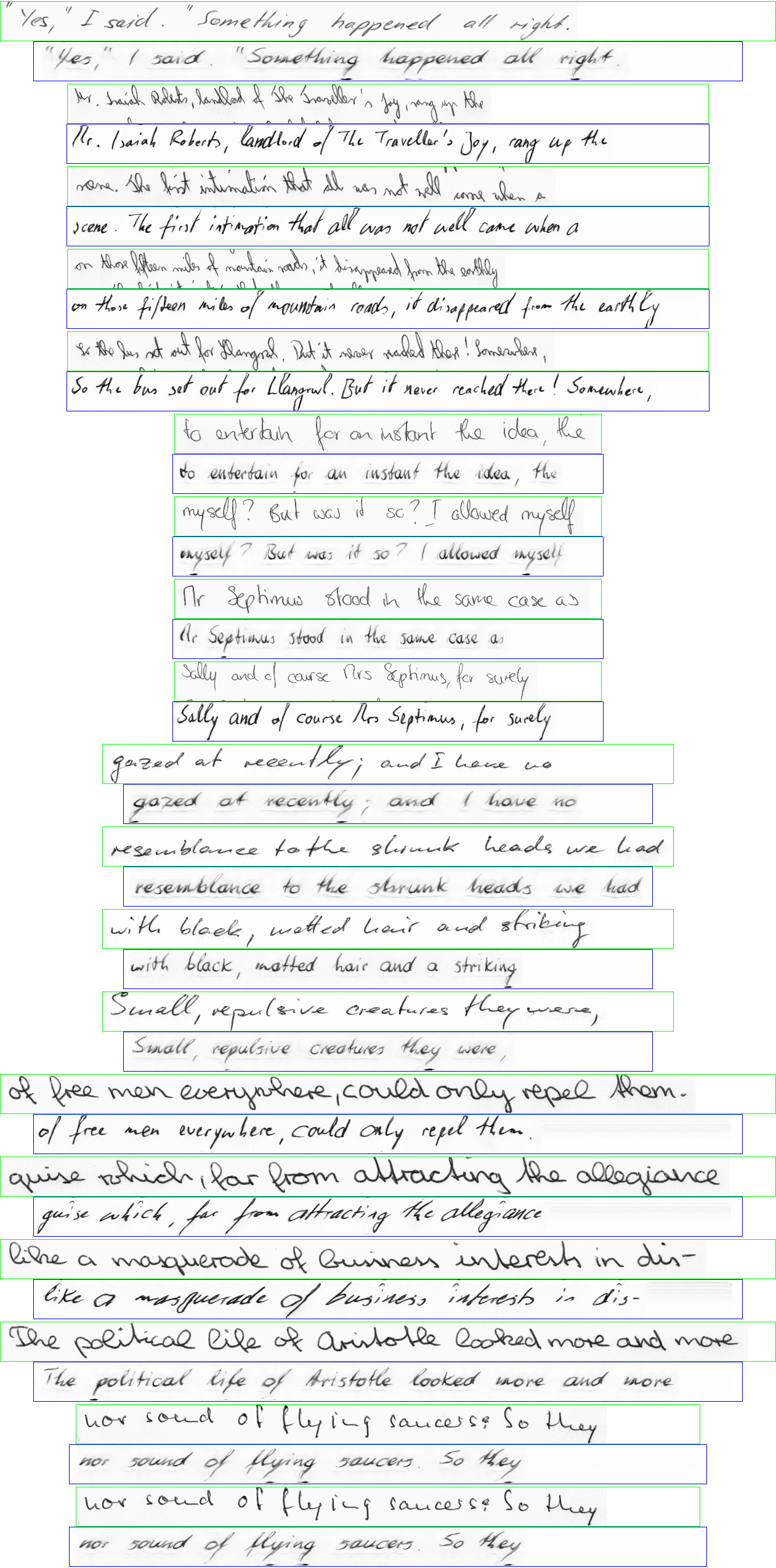}
\caption{Additional ablation results, without character specific components of $S$.}
\label{fig:supp_noCharSpec}
\end{figure}

\begin{figure}[h]
\centering
\includegraphics[width=0.72\textwidth]{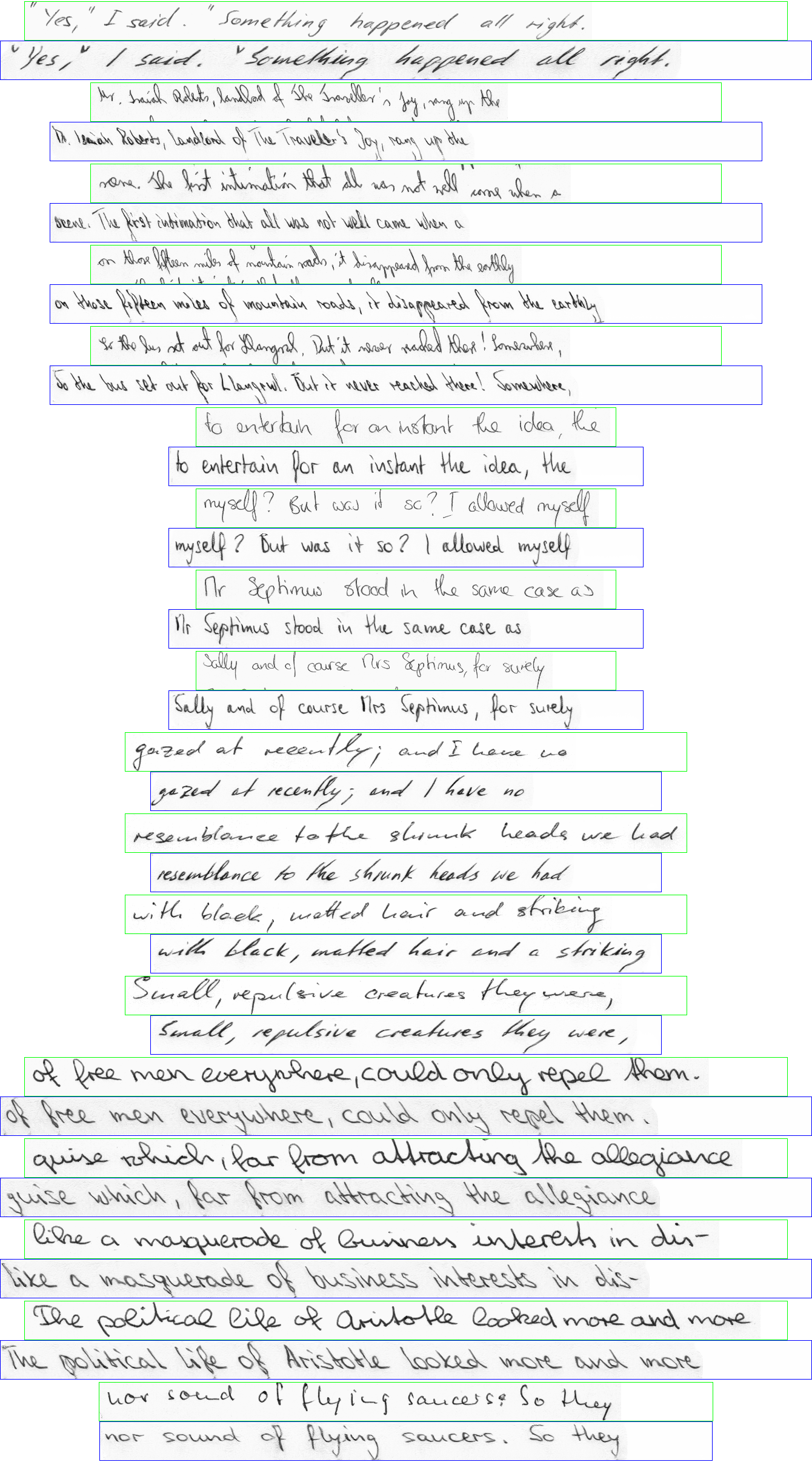}
\caption{Additional ablation results, without pixel-wise reconstruction loss.}
\label{fig:supp_noPix}
\end{figure}

We present additional results for each of the ablation models:
\begin{itemize}
    \item Fig.~\ref{fig:supp_noAuto}: No reconstruction loss
    \item Fig.~\ref{fig:supp_noGAN}: No adversarial loss
    \item Fig.~\ref{fig:supp_noHWR}: No handwriting recognition supervision
    \item Fig.~\ref{fig:supp_noCharSpec}: No character specific components of $S$
    \item Fig.~\ref{fig:supp_noPix}: No pixel reconstruction loss
\end{itemize}

\clearpage

\subsection{Model Specifications}\label{sec:model_supp}

\begin{figure}[b]
    \centering
    \begin{minipage}{0.48\textwidth}
        \centering
        \includegraphics[width=0.99\textwidth]{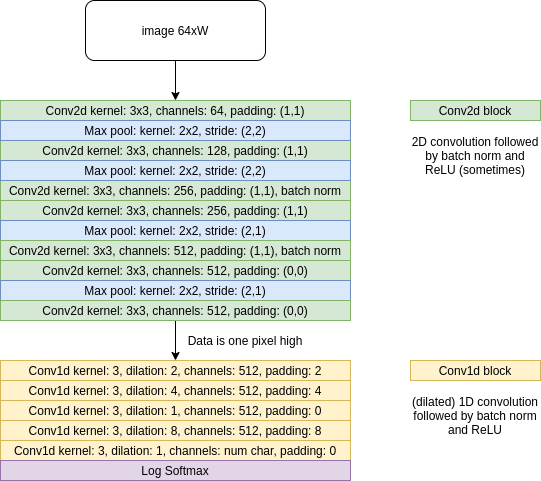} 
        \caption{Handwriting recognition network $R$ architecture}
\label{fig:hwr}
    \end{minipage}\hfill
    \begin{minipage}{0.48\textwidth}
        \centering
        \includegraphics[width=0.99\textwidth]{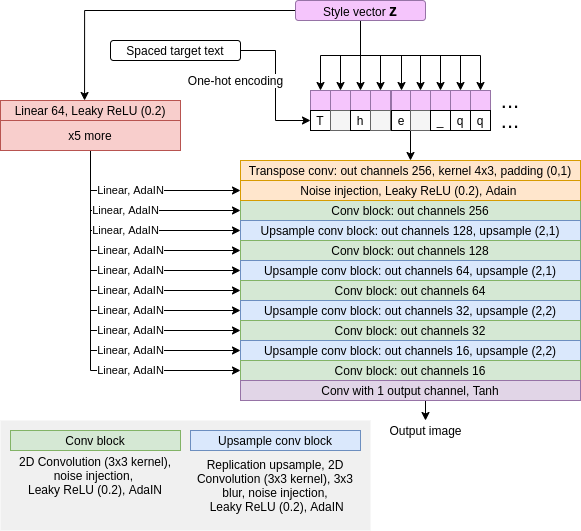} 
        \caption{Generator $G$ architecture}
\label{fig:generator}
    \end{minipage}
\end{figure}

\begin{figure}
    \centering
    \begin{minipage}{0.48\textwidth}
        \centering
        \includegraphics[width=0.99\textwidth]{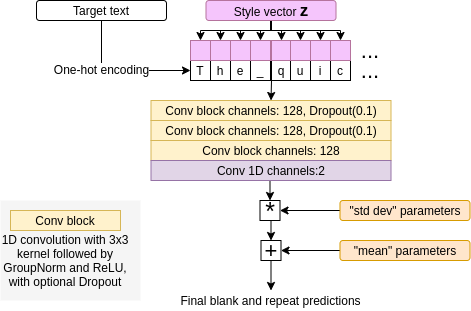} 
        \caption{Spacer network $C$ which predicts the spaced text. It predicts the number of blanks proceeding each character and the number of times the character should be repeated.
}
\label{fig:spacer}
    \end{minipage}\hfill
    \begin{minipage}{0.48\textwidth}
        \centering
        \includegraphics[width=0.99\textwidth]{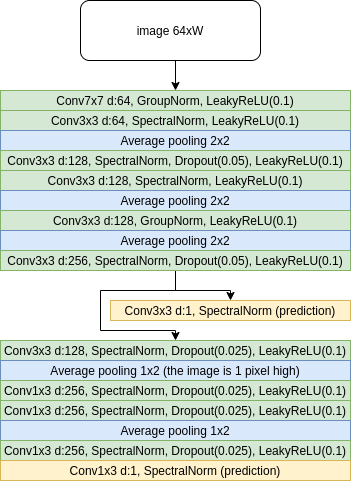} 
        \caption{Discriminator $D$ architecture. 
}
\label{fig:disc}
    \end{minipage}
\end{figure}

\begin{figure}[]
\centering
\includegraphics[width=0.95\textwidth]{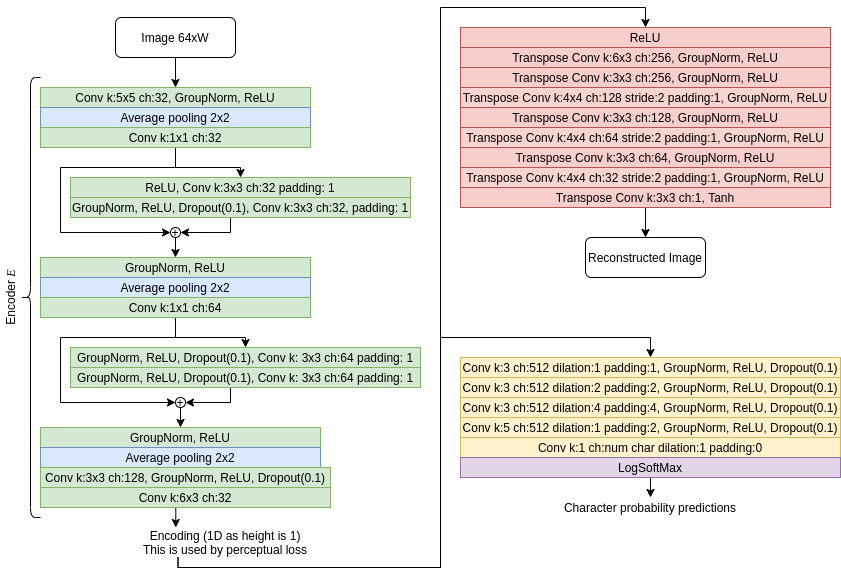}
\caption{Encoder network $E$ (green) and auxiliary decoder (red) and recognition head (yellow) used to train $E$.}
\label{fig:encoder}
\end{figure}

\begin{figure}[]
\centering
\includegraphics[width=0.95\textwidth]{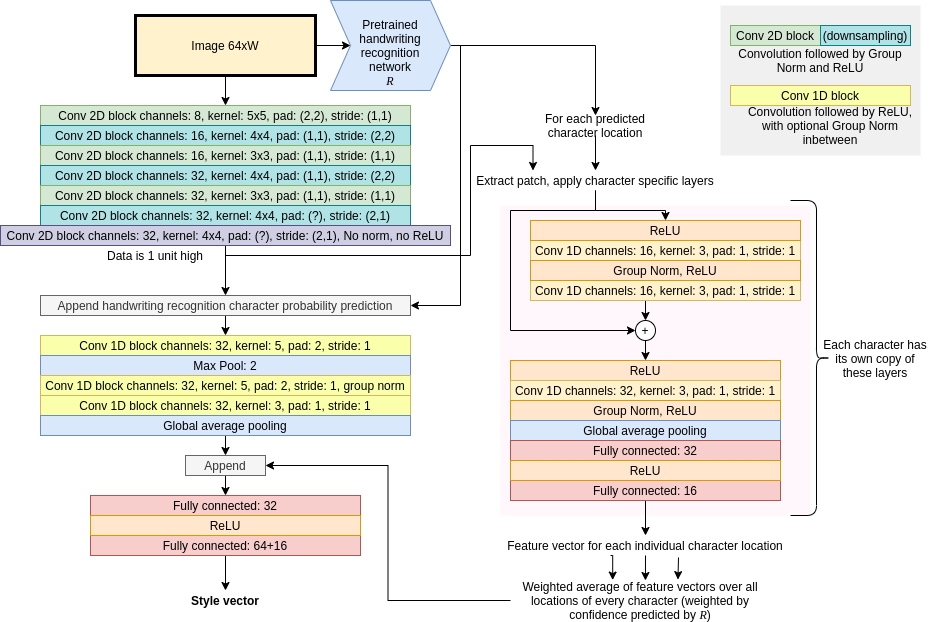}
\caption{Style Extractor $S$. It leverages the output of $R$ both as additional input and to (roughly) locate characters. The locations are used to crop features to pass to character specific layers (the learn to extract features for one character).}
\label{fig:supp_style_extractor}
\end{figure}

We present here detailed diagrams of various parts of the model:
\begin{itemize}
    \item Fig.~\ref{fig:hwr}: The handwriting recognition model $R$
    \item Fig.~\ref{fig:generator}: The generator $G$
    \item Fig.~\ref{fig:spacer}: The auxiliary spacing network $C$
    \item Fig.~\ref{fig:disc}: The discriminator $D$
    \item Fig.~\ref{fig:encoder}: The encoder $E$
    \item Fig.~\ref{fig:supp_style_extractor}: The style extractor $S$
\end{itemize}

The encoder $E$ is trained using the same IAM training set. It is jointly trained with a decoder as an autoencoder with an L1 reconstruction loss and as a  handwriting recognition network with a recognition head using the CTC loss. It is trained with the Adam optimizer 6000 iterations with a learning rate of 0.0002.

\end{document}